\pdfoutput=1

\documentclass[11pt]{article}

\usepackage[]{acl}

\pdfoutput=1
\usepackage{times}
\usepackage{latexsym}
\usepackage{color}
\usepackage{pgfplots}
\pgfplotsset{width=10cm,compat=1.16}
\usepackage{subcaption}
\usepackage{bm} 
\usepackage{amsmath}
\usepackage{listings}
\usepackage{adjustbox}
\usepackage{pgf-pie}
\expandafter\def\expandafter\normalsize\expandafter{%
    \normalsize%
    \setlength\abovedisplayskip{-5pt}%
    \setlength\belowdisplayskip{-9pt}%
    \setlength\abovedisplayshortskip{0pt}%
    \setlength\belowdisplayshortskip{0pt}%
}

\usepackage{mathtools}
\usepackage{algorithm,algorithmic}
\usepackage{cases}

\usepackage[T1]{fontenc}

\usepackage[utf8]{inputenc}

\usepackage{microtype}
\usepackage{url}

\usepackage{multirow}
\usepackage{tabularx}
\usepackage{lipsum} 
\usepackage{booktabs}
\usepackage{inconsolata}
\usepackage{graphicx}
\usepackage{calligra}
\usetikzlibrary{pgfplots.polar}

\definecolor{color1}{RGB}{159,168,218}
\definecolor{color2}{RGB}{173,201,198}
\definecolor{color3}{RGB}{207,216,220}
\definecolor{color4}{RGB}{247,194,183}
\definecolor{color5}{RGB}{187,230,190}
\definecolor{color6}{RGB}{187,230,222}
\definecolor{darkgreen}{rgb}{0,0.5,0}
\definecolor{fancy_color1}{RGB}{51,75,127}
\definecolor{fancy_color2}{RGB}{180,75,144}
\definecolor{pie_pink}{HTML}{feaba7}
\definecolor{pie_rose}{HTML}{fbe1ed}
\definecolor{pie_purple}{HTML}{dbd1ed}
\definecolor{pie_green}{HTML}{adeaad}
\definecolor{pie_orange}{HTML}{ffcd88}
\definecolor{pie_gray}{HTML}{dde7eb}
\definecolor{orchid}{RGB}{218, 112, 214}

\definecolor{radar_color1}{HTML}{0077ad}
\definecolor{radar_color2}{HTML}{aa71c7}
\definecolor{radar_color3}{HTML}{ff7f4f}
\definecolor{radar_color4}{HTML}{e765af}
\definecolor{radar_color5}{HTML}{6178c5}
\definecolor{radar_color6}{HTML}{ffa600}
\definecolor{radar_color7}{HTML}{ff6684}

\definecolor{bar_color1}{HTML}{0099de}
\definecolor{bar_color2}{HTML}{7a92ee}
\definecolor{bar_color3}{HTML}{c684e5}
\definecolor{bar_color4}{HTML}{ff72c2}
\definecolor{bar_color5}{HTML}{ff6e8d}
\definecolor{bar_color6}{HTML}{ff8453}
\definecolor{bar_color7}{HTML}{ffa600}

\title{Graphusion: Leveraging Large Language Models for \\Scientific Knowledge Graph Fusion and Construction in NLP Education}

\author{
Rui Yang$^{1*}$, Boming Yang$^2$, Sixun Ouyang$^3$, Tianwei She$^3$, Aosong Feng$^4$, \\ 
\textbf{Yuang Jiang$^3$, Freddy Lecue$^5$, Jinghui Lu$^3$, Irene Li$^{2,3*}$} \\ 
\\
$^1$Duke-NUS Medical School,  
$^2$University of Tokyo,  
$^3$Smartor Inc., \\
$^4$Yale University, 
$^5$INRIA \\
yang.rui@duke-nus.edu, ireneli@ds.itc.u-tokyo.ac.jp
}

\begin{document}
\maketitle
\begin{abstract}
Knowledge graphs (KGs) are crucial in the field of artificial intelligence and are widely applied in downstream tasks, such as enhancing Question Answering (QA) systems. The construction of KGs typically requires significant effort from domain experts. Recently, Large Language Models (LLMs) have been used for knowledge graph construction (KGC), however, most existing approaches focus on a local perspective, extracting knowledge triplets from individual sentences or documents. In this work, we introduce Graphusion, a zero-shot KGC framework from free text. The core fusion module provides a global view of triplets, incorporating entity merging, conflict resolution, and novel triplet discovery. We showcase how Graphusion could be applied to the natural language processing (NLP) domain and validate it in the educational scenario. Specifically, we introduce TutorQA, a new expert-verified benchmark for graph reasoning and QA, comprising six tasks and a total of 1,200 QA pairs. Our evaluation demonstrates that Graphusion surpasses supervised baselines by up to 10\% in accuracy on link prediction. Additionally, it achieves average scores of 2.92 and 2.37 out of 3 in human evaluations for concept entity extraction and relation recognition, respectively. 


\end{abstract}

\section{Introduction}


Recently, large language models (LLMs) such as GPT ~\cite{achiam2023gpt} and LLaMA~\cite{touvron2023llama} have demonstrated outstanding performance across various tasks in the field of natural language processing (NLP) ~\cite{yang2023large, yang2023ascle, song2023nlpbenchevaluatinglargelanguage, yang2023going, gao2024largelanguagemodelswikipediastyle}. However, the content generated by LLMs often lacks accuracy and interpretability~\cite{zhang2023siren, yang2024retrieval}. To address these challenges, one approach is leveraging Knowledge Graphs (KGs) to enhance LLMs ~\cite{Yang2023ChatGPTIN, yang2024kg}. By prompting the structured KG knowledge to LLMs, they can generate more reliable content. Additionally, KG-enhanced LLMs can be applied to various KG tasks as well, including graph completion, reasoning and more~\cite{Zhu2023LLMsFK, Chen2023IncorporatingSS}.


\begin{figure}[t]
    \centering
    \includegraphics[width=1.00\linewidth]{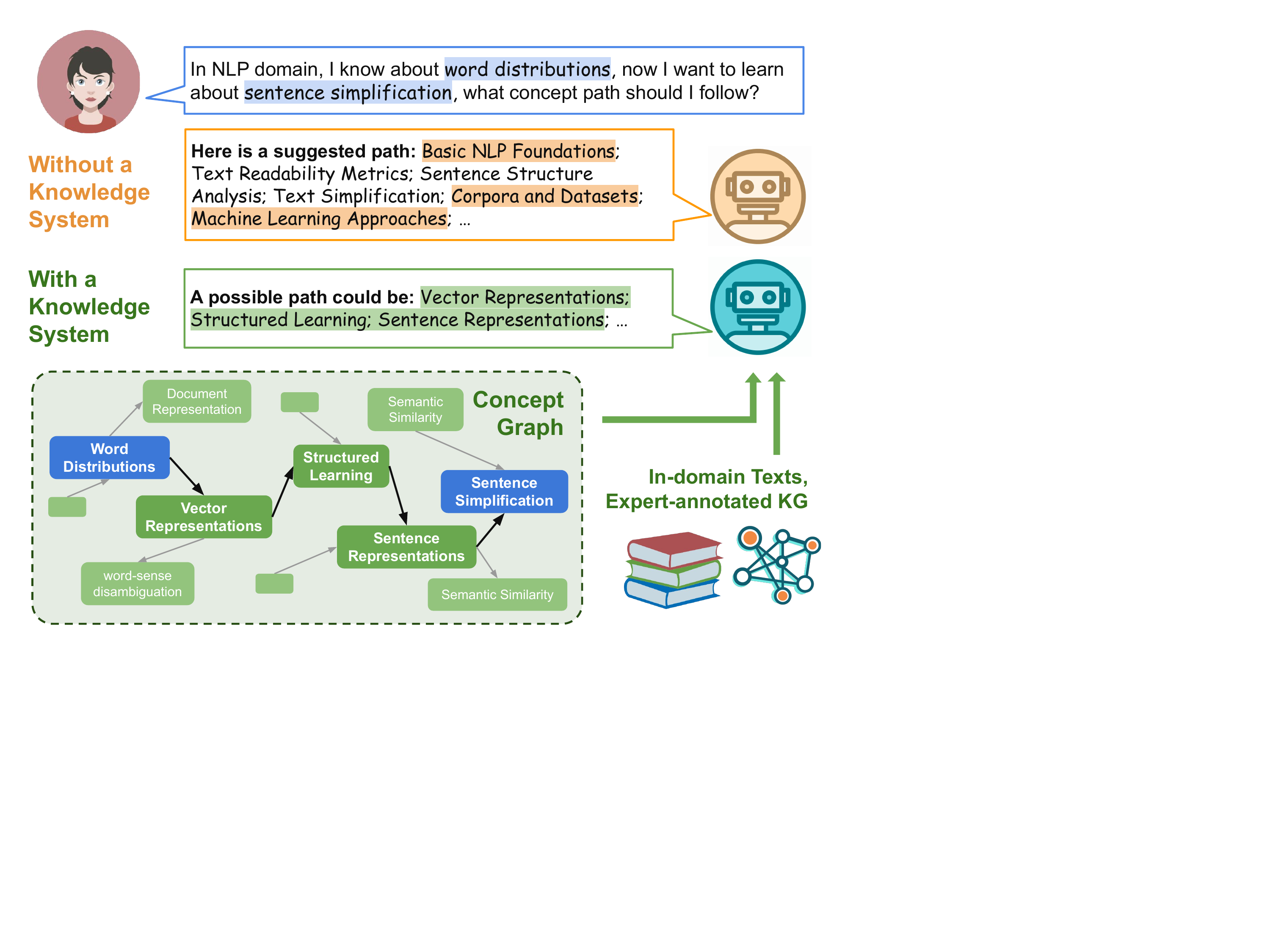}
    \caption{Comparison of QA systems with and without a knowledge system.}
    \label{fig:intro}
    \vspace{-6mm}
\end{figure}

In Fig ~\ref{fig:intro}, we demonstrate an example in the educational scenario. A user asks a question involving specific concepts (highlighted in blue). Ideally, the response should reflect the relations between these concepts, essentially outlining the learning path that connects them. However, without a knowledge system, an LLM might offer answers that are somewhat relevant but too general, including broad concepts like "Basic NLP Foundations", or it may introduce confusing concepts with inaccurate specificity ("Corpora and Datasets"). In contrast, when equipped with a knowledge system (such as a concept graph showing prerequisite relations) and supplemented with relevant in-domain texts, the response becomes more refined, reflecting a deeper understanding of the concept relations.



Automatic methods have been applied to knowledge graph construction (KGC)~\cite{Sheng2022ChallengingTA, Baek2023KnowledgeAugmentedLM, Carta2023IterativeZL}, with most of them employing a localized perspective, extracting triplets at the sentence or paragraph level, which is suitable for shallow knowledge, such as (\texttt{people, belong\_to, organization}). However, this localized approach often fails to capture the comprehensive and interconnected nature of knowledge. The accuracy and completeness of triplets can be significantly limited when sourced from isolated segments of text, which is essential for scientific graphs containing deep knowledge.

Recognizing this limitation, we propose a significant paradigm shift towards a global view in KGC. Our approach includes a graph fusion module that extracts candidate triplets and performs global merging and resolution across multiple sources. Specifically, we leverage LLMs not only for extraction but also for critical knowledge integration, marking the first initiative to utilize LLMs for such a comprehensive merging process. We believe that this global perspective is crucial for constructing more accurate and holistic KGs, as it allows for the consideration of broader contexts and relations that span beyond single documents. Similarly, this is particularly vital in scientific KGs, where the relations between complex concepts cannot be adequately understood by examining individual sentences.

In this study, we utilize LLMs to construct and fuse scientific KGs, focusing primarily on the domain of natural language processing. Most importantly, we apply the constructed KG in the educational question-answering (QA) scenario. Our contributions are summarized as follows. First, we propose the Graphusion framework, which allows zero-shot KGC from free text. The core graph fusion component incorporates entity merging, conflict resolution, and novel triplet discovery. Evaluation results show that Graphusion achieves scores of 2.92 and 2.37 out of 3 for entity extraction and relation recognition, respectively, demonstrating its potential for automatic and large-scale KGC. To the best of our knowledge, our work is pioneering in scientific KGC with fusion using the zero-shot capabilities of LLMs. Second, we present TutorQA, a QA benchmark featuring six diverse tasks and comprising 1,200 expert-verified, NLP-centric QA pairs designed to mimic college-level tutoring questions. Third, we develop a pipeline to enhance the interaction between LLMs and the concept graph for TutorQA, achieving significant results across all tasks. All the code and data can be found in \url{https://github.com/IreneZihuiLi/CGPrompt}.

\section{Related Work}


\paragraph{Knowledge Graph Construction} KGC aims to create a structured representation of knowledge in the form of a KG. Research on KGs spans various domains, including medical, legal, and more ~\cite{wikidata, Li2020RVGAERG, LeTuan2022TowardsBL, Kalla2023ScientificKG, ahrabian2023pubgraph}. Typically, KGC involves several methods such as entity extraction and link prediction ~\cite{Luan2018MultiTaskIO,Reese2020KGCOVID19AF}, with a significant focus on supervised learning. Recently, LLMs have been utilized in KGC relying on their powerful zero-shot capabilities ~\cite{Zhu2023LLMsFK, Carta2023IterativeZL}. Although relevant works propose pipelines for extracting knowledge, they often remain limited to localized extraction, such as at the sentence or paragraph level. In contrast, our work focuses on shifting from a local perspective to a global one, aiming to generate a more comprehensive KG.



\paragraph{Educational NLP} Modern NLP and Artificial Intelligence (AI) techniques have been applied to a wide range of applications, with education being a significant area. For instance, various tools have been developed focusing on writing assistance, language study, automatic grading, and quiz generation~\cite{Seyler2015GeneratingQQ, GonzlezCarrillo2021AutomaticGT, Zhang2023VISARAH,Lu2023ErrorAP}. Moreover, in educational scenarios, providing responses to students still requires considerable effort, as the questions often demand a high degree of relevance to the study materials and strong domain knowledge. Consequently, many studies have concentrated on developing automatic QA models~\cite{Zylich2020ExploringAQ, Hicke2023ChaTATA}, which tackle a range of queries, from logistical to knowledge-based questions. In this work, we integrate an LLM-constructed KG for various QA tasks in NLP education.


\section{Graphusion Framework}


\begin{figure*}[t!]
    \centering
    \makebox[\textwidth][c]{
        \includegraphics[width=1.05\textwidth]{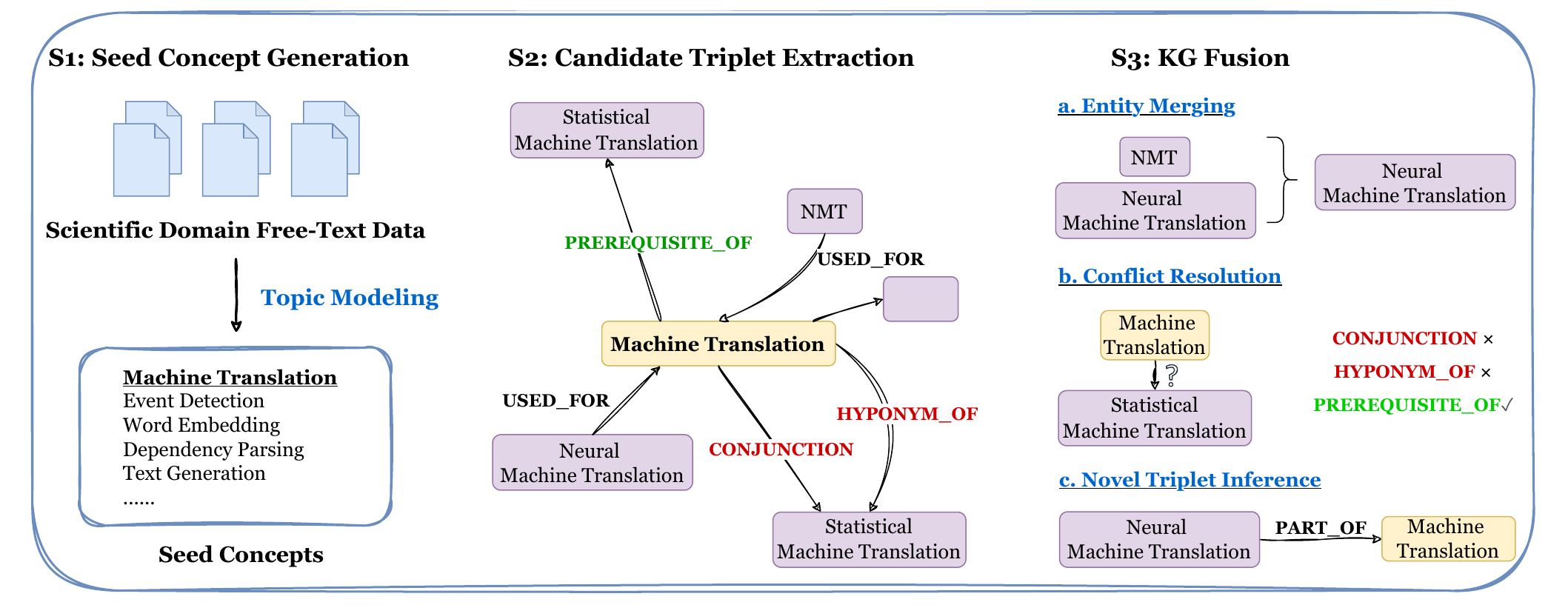}
    }
    \caption{Graphusion framework illustration. Gaphusion consists of 3 steps: S1 Seed Concept Generation, S2 Candidate Triplet Extraction and S3 KG Fusion.}
    \label{fig:model}
    \vspace{-6mm}
\end{figure*}

In this section, we introduce our Graphusion Framework for scientific KGC. 

\subsection{Problem Definition}

A $KG$ is defined as a set of triplets: $KG = \{(h_i, r_i, t_i) \mid h_i, t_i \in E, r_i \in R, i = 1, 2, \ldots, n\} $,
where: $E$ is the set of entities, $R$ is the set of possible relations, and $n$ is the total number of triplets in the $KG$. The task of zero-shot KGC involves taking a set of free text $T$ and generating a list of triplets $(h,r,t)$. Optionally, there is an expert-annotated KG $EG$ as input, in order to provide some existing knowledge. In our setting, the number of triplets of $KG$ is much larger than $EG$. We select the domain to be NLP, so the entities are concepts only, other entity types such as people, organizations are not our focus. Following previous works \cite{Luan2018MultiTaskIO}, we define 7 relations types: \texttt{Prerequisite\_of}, \texttt{Used\_for}, \texttt{Compare}, \texttt{Conjunction}, \texttt{Hyponym\_of}, \texttt{Evaluate\_for} and \texttt{Part\_OF}.


\subsection{Zero-shot Link Prediction}

While the task of KGC is to generate a list of triplets, including entities and their corresponding relations, we start with a simpler setting: focusing solely on link prediction for pre-defined entity pairs and a single relation type. This setting helps us understand the capabilities of LLMs on scientific KGC under a zero-shot setting. Specifically, given a concept pair $(A, B)$, the task of link prediction is to determine if a relationship $r$ exists. We choose $r=$\texttt{Prerequisite\_of}. For instance, the relation "Viterbi Algorithm" $\rightarrow$ "POS Tagging" implies that to learn the concept of "POS Tagging," one must first understand the "Viterbi Algorithm." Initially, a predefined set of concepts $\calligra{C}$ is given.

\paragraph{LP Prompt} We then design a Link Prediction (LP) Prompt to solve the task. The core part is to provide the domain name, the definition and description of the dependency relation to be predicted, and the query concepts. Meanwhile, we explore whether additional information, such as concept definitions from Wikipedia and neighboring concepts from training data (when available), would be beneficial. The LP Prompt is as follows:

\begin{lstlisting}[language={}, captionpos=b, label={lst: lp}, basicstyle=\linespread{0.85}\ttfamily, frame=single, breaklines=true, keepspaces=true]
We have two {domain} related concepts: 
A: {concept_1} and B: {concept_2}.

Do you think learning {concept_1} 
will help in understanding {concept_2}?

Hints:
1. Answer YES or NO only.
2. This is a directional relation, 
   which means if the answer is "YES", 
   (B, A) is false, but (A, B) is true.
3. Your answer will be used to create a 
   knowledge graph.
   
{Additional Information}
\end{lstlisting}
\vspace{-3mm}

\subsection{Graphusion: Zero-shot Knowledge Graph Construction}
\label{sec:kgc}
We now introduce our Graphusion framework for constructing scientific KGs, shown in Fig ~\ref{fig:model}. Our approach addresses three key challenges in zero-shot KGC: 1) the input consists of free text rather than a predefined list of concepts; 2) the relations encompass multiple types, and conflicts may exist among them; and 3) the output is not a single binary label but a list of triplets, making evaluation more challenging.

\textbf{Step 1: Seed Concept Generation}
Extracting domain-specific concepts using LLMs under a zero-shot setting is highly challenging due to the absence of predefined concept lists. This process is not only resource-intensive but also tends to generate a large number of irrelevant concepts, thereby compromising the quality of extraction. To address these issues, we adopt a seed concept generation approach for efficiently extracting in-domain concepts from free text \cite{ke2024comparing}. Specifically, we utilize BERTopic \cite{Grootendorst2022BERTopicNT} for topic modeling to identify representative concepts for each topic. These representative concepts serve as seed concepts, denoted as $Q$. The initialized seed concepts ensure high relevance in concept extraction and provide certain precision for subsequent triplet extraction.

\textbf{Step 2: Candidate Triplet Extraction}
Based on these seed concepts, in Step 2, we begin extracting candidate triplets from the free text. Each time, we input a concept \( q \in Q \) (\{query\}) as the query concept and retrieve documents containing this concept (\{context\}) through information retrieval. Our goal is to extract any potential triplets that include this query concept. To achieve this, we design a Chain-of-Thought (CoT) \cite{CoT} prompt. We first instruct the LLMs to extract in-domain concepts, then identify the possible relations between those concepts and $q$. Then, we ask LLMs to discover novel triplets, even if $q$ is not initially included. This approach ensures that the seed concepts play a leading role in guiding the extraction of in-domain concepts. Meanwhile, the candidate triplets will encompass novel concepts. We design the \textbf{Extraction Prompt} to be the following:

\begin{lstlisting}[language={}, captionpos=b, label={lst: zs}, basicstyle=\linespread{0.85}\ttfamily, frame=single, breaklines=true, keepspaces=true]
Given a context {context}, and a query 
concept {query}, do the following: 

1. Extract the query concept and 
   in-domain concepts from the context, 
   which should be fine-grained...
2. Determine the relations between 
   the query concept and the extracted 
   concepts, in a triplet format: 
   (<head concept>, <relation>, <tail concept>)...
   {Relation Definition}
3. Please note some relations are 
   strictly directional...
4. You can also extract triplets from 
   the extracted concepts, and the 
   query concept may not be necessary 
   in the triplets. 
\end{lstlisting}
\vspace{-1mm}

After processing all the queries from the seed concept list, we save all the candidate triplets. We denote this zero-shot constructed KG by the LLM as $\mathcal{ZS-KG}$.

\textbf{Step3: Knowledge Graph Fusion} The triplets extracted in the previous step provide a local view rather than a global perspective of each queried concept. Due to the limitations of context length, achieving a global view is challenging. Additionally, the relations extracted between two concepts can sometimes be conflicting, diverse, or incorrect, such as \textbf{(neural summarization methods, Used-for, abstractive summarization)} and \textbf{(neural summarization methods, Hyponym-of, abstractive summarization)}. To address the aforementioned challenge, we propose the Fusion step. This approach helps reconcile conflicting relations, integrate diverse or incorrect relations effectively, and ultimately provides a global understanding of an entity pair.


Specifically, for each query concept $q$, we first query from $\mathcal{ZS-KG}$, and obtain a sub-graph that contains $q$:

\vspace{-1mm}
\begin{equation}
\text{LLM-KG} = \{ (h, r, t) \in \text{ZS-KG}
\mid h = q \text{ or } t = q \}.
\nonumber
\end{equation}
\vspace{-1mm}

Optionally, if there is an expert-annotated KG available, we will also query a sub-graph, marked as $\mathcal{E-G}$. Moreover, we conduct a dynamic retrieval of $q$ again from the free text (\{background\}), to help LLMs to have a better understanding on how to resolve the conflicted triplets. This key fusion step focuses on three parts: a) entity merging: merge semantically similar entities, i.e., \texttt{NMT} vs \texttt{neural machine translation}; b) conflict resolution: for each entity pair, resolve any conflicts and choose the best one; and c) novel triplet inference: propose new triplet from the background text. We utilize the following \textbf{Fusion Prompt}:

\begin{lstlisting}[language={}, captionpos=b, label={lst: fusion}, basicstyle=\linespread{0.85}\ttfamily, frame=single, breaklines=true, keepspaces=true]
Please fuse two sub-knowledge graphs 
about the concept "{concept}".
Graph 1: {LLM-KG}   Graph 2: {E-G}

Rules for Fusing the Graphs:
1. Union the concepts and edges. 
2. If two concepts are similar, or refer 
   to the same concept, merge them into 
   one concept, keeping the one that is 
   meaningful or specific ...
3. Only one relation is allowed between 
   two concepts. If there is a conflict, 
   read the "### Background" to help you 
   keep the correct relation...
4. Once step 3 is done, consider every 
   possible concept pair not covered in
   step 2. For example, take a concept 
   from Graph 1, and match it with a 
   concept from Graph 2. Then, please
   refer to "### Background" to summarize
   new triplets. 

### Background: 
{background}

{Relation Definition}
\end{lstlisting}

\subsection{Evaluation}

The evaluation of KGC is challenging since each model generating different triplets from the free text, along with the lack of expert annotations. To address this, we conduct the expert evaluation on the pipeline output. We ask experts to assess both \textit{concept entity quality} and \textit{relation quality}, providing ratings ranging from 1 to 3. The former measures the relevance and specificity of the extracted concepts, while the latter evaluates the logical accuracy between concepts. Additionally, we calculate the Inter-Annotator Agreement (IAA) of the two experts' evaluations using the Kappa score.

\section{Experiments}

\subsection{Link Prediction}
We conduct experiments using the LectureBankCD dataset~\cite{Li2021UnsupervisedCP} and report the performance on the NLP domain. LectureBankCD contains up to 322 pre-defined NLP concepts and prerequisite relation labels on the concept pairs. We benchmark on the official test set against the following \textbf{Supervised Baselines}: DeepWalk~\cite{Perozzi2014DeepWalkOL}, and Node2vec~\cite{Grover2016node2vecSF}, P2V~\cite{Wu2020Phrase2VecPE}, and BERT~\cite{Devlin2019BERTPO}.  These methods utilize pre-trained or graph-based models to encode concept embeddings and then perform binary classification to determine the presence of positive or negative edges in given concept pairs. In our LLM-based experiments, we show two main settings: \textbf{Zero-shot}, which employs LP Prompt; and \textbf{Zero-shot with RAG}, which enhances zero-shot with the addition of Retrieval Augmented Generation (RAG) method~\cite{krishna2023prompt}. RAG has shown to improve on existing LLMs on text generation tasks such as QA. In Tab. ~\ref{tab:nlp_res_LP}, we observe that the zero-shot performance of GPT-4 and GPT-4o surpasses that of the best traditional supervised baseline. This suggests that LLMs can recover a domain-specific concept graph without relying on expert annotations. With the aid of RAG, which incorporates more domain-specific data, GPT-4o achieves significant improvements. 

\begin{table}[ht]
\centering
\small
\begin{tabular}{lcc}
\toprule
\textbf{Model} & \textbf{Accuracy} & \textbf{F1 Score} \\
\midrule
\multicolumn{3}{l}{\textit{Supervised Baselines}} \\
DeepWalk & 0.6292 & 0.5860 \\
P2V & 0.6369 & 0.5961 \\
Node2vec & 0.6209 & 0.6181 \\
BERT & \underline{0.7088} & \underline{0.6963} \\
\midrule
\multicolumn{3}{l}{\textit{Zero-shot}} \\
LLaMa2 & 0.6058 & 0.6937 \\
GPT-3.5 & 0.6123 & 0.7139 \\
GPT-4 & 0.7639 & 0.7964 \\
GPT-4o & 0.7980 & 0.7958 \\
\midrule
\multicolumn{3}{l}{\textit{Zero-shot with RAG}}  \\
GPT-3.5 & 0.7587 & 0.7793 \\
GPT-4 & 0.7755 & 0.7958 \\
GPT-4o & \textbf{0.8117} & \textbf{0.8181} \\
\hline
\end{tabular}
\caption{Evaluation of the link prediction task on the LectureBankCD-NLP test set.}
\label{tab:nlp_res_LP}
\vspace{-6mm}
\end{table}



\subsection{Knowledge Graph Completion}

To conduct KGC, we need a large-scale free-text corpus to serve as the knowledge source. Since there is no standard benchmark for evaluation, we collected the proceedings papers from the ACL conference\footnote{\url{https://aclanthology.org/venues/acl/}} spanning 2017-2023, which includes a total of 4,605 papers. Considering that abstracts provide high-density, noise-free information and save computational resources, we perform topic modeling on the abstracts of these articles. Eventually, we successfully generate 688 seed concepts. 
We implement Graphusion on four selected models: LLaMa3-70b\footnote{\url{https://llama.meta.com/llama3/}}, GPT-3.5, GPT-4 and GPT-4o. Two domain experts participate in the evaluation. Tab. ~\ref{tab:IAA} shows the evaluation results of these models on the quality of concept entity and relation, as well as the experts' consistency score. Overall, the rating for concept entity surpasses relation, demonstrating the challenging of relation extraction. 
Among all, GPT-4o achieves the best on both concept and relation. 
Additionally, the high consistency score among the experts indicates the reliability of the expert evaluation.

\begin{table}[ht]
\centering
\small
\begin{tabular}{lccccc}
\toprule
\textbf{Model} & \multicolumn{2}{c}{\textbf{Concept Entity}} & \multicolumn{2}{c}{\textbf{Relation}} \\
\cmidrule(r){2-3} \cmidrule(r){4-5}
& \textbf{Rating} & \textbf{Kappa} & \textbf{Rating} & \textbf{Kappa} \\
\midrule
LLaMA & $2.83_{+0.47}$ & 0.63 & $1.82_{+0.81}$ & 0.51 \\
GPT-3.5 & $2.90_{+0.38}$ & 0.48 & $2.14_{+0.83}$ & 0.67 \\
GPT-4 & $2.84_{+0.50}$ & 0.68 & $2.36_{+0.81}$ & 0.65 \\
GPT-4o & $\bm{2.92_{+0.32}}$ & 0.65 & $\bm{2.37_{+0.82}} $ & 0.67 \\
\bottomrule
\end{tabular}
\caption{Rating for the qaulity of concept entity and relation, and IAA score for the expert evaluation.}
\label{tab:IAA}
\vspace{-6mm}
\end{table}


\begin{table*}[t]
\centering
\small
\begin{tabular}{lcccc}
\toprule
\textbf{Dataset} & \textbf{Domain} & \textbf{Answer Type} & \textbf{With KG} & \textbf{Collection}\\ \midrule
CBT~\cite{Hill2015TheGP} & Open-Domain & Multiple Choice & No & Automated \\
LectureBankCD~\cite{Li2021UnsupervisedCP} &  NLP, CV, BIO & Binary & Yes & Expert-verified \\
FairytaleQA~\cite{Xu2022FantasticQA} & Open-Domain & Open-ended & No & Expert-verified \\
ChaTa~\cite{Hicke2023ChaTATA} & CS & Free Text & No & Students \\ 
ExpertQA~\cite{Malaviya2023ExpertQAEQ} & Science & Free Text & No & Expert-verified \\
\midrule
TutorQA (this work) & NLP & Open-Ended, Entity List, Binary & Yes & Expert-verified \\ 
\bottomrule
\end{tabular}
\caption{Comparison with other similar benchmarks: Educational or general QA benchmarks. }
\label{tab:tabqa}
\vspace{-4mm}
\end{table*}

\section{TutorQA}
We introduce the TutorQA benchmark, a QA dataset designed for concept graph reasoning and text generation in the NLP domain. TutorQA comprises six categories, encompassing a total of 1,200 QA pairs that have been validated by human experts. These questions go beyond simple syllabus inquiries, encompassing more extensive and challenging topics that require interaction with the completed graph, as well as proficiency in text comprehension and question answering. We list some similar benchmarks in Tab.~\ref{tab:tabqa}. While numerous open-domain QA benchmarks exist, our focus has been primarily on those within the scientific domain and tailored for college-level education, aligning with our objective to compare with benchmarks that can emulate a learning scenario. Among those, TutorQA is distinguished by its diversity in answer types and features expert-verified questions, ensuring a high standard of quality and relevance.

\subsection{QA Tasks}
We summarize the tasks and provide example data in Fig~\ref{fig:tutorqa_sample}. More data statistics and information can be found in Appendix~\ref{app:tutorqa_info}. 

\begin{figure*}[t!]
    \centering
\includegraphics[width=1.00\textwidth]{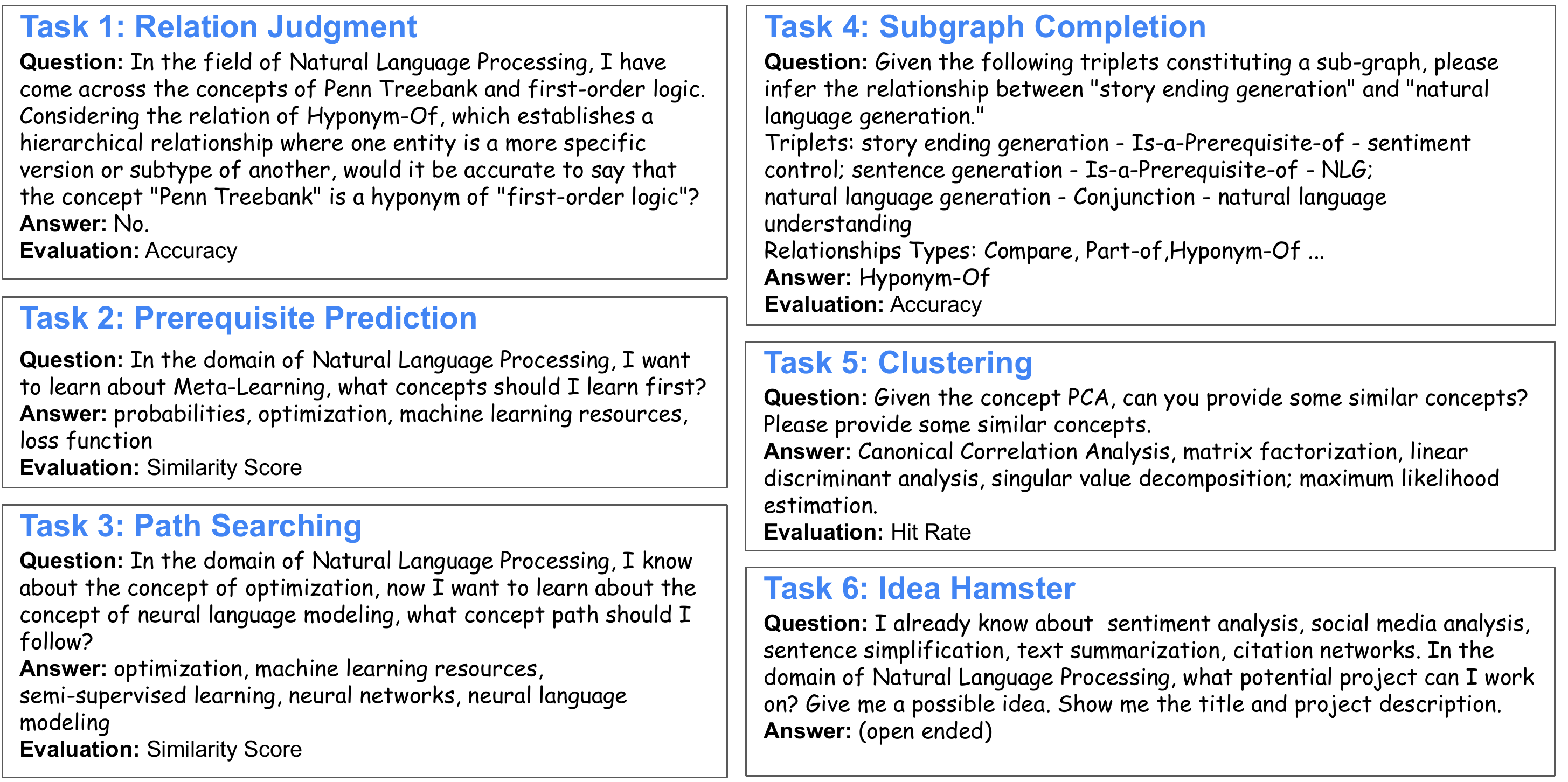}
    \caption{TutorQA tasks: We present a sample data instance and the corresponding evaluation metric for each task. Note: Task 6 involves open-ended answers, which are evaluated through human assessment. }
    \label{fig:tutorqa_sample}
    \vspace{-4mm}
\end{figure*}

\textbf{Task 1: Relation Judgment} 
The task is to assess whether a given triplet, which connects two concepts with a relation, is accurate.

\textbf{Task 2: Prerequisite Prediction} 
The task helps students by mapping out the key concepts they need to learn first to understand a complex target topic.

\textbf{Task 3: Path Searching} %
This task helps students identify a sequence of intermediary concepts needed to understand a new target concept by charting a path from the graph.

\textbf{Task 4: Subgraph Completion} 
The task involves expanding the KG by identifying hidden associations between concepts in a subgraph.

\textbf{Task 5: Similar Concepts} 
The task requires identifying concepts linked to a central idea to deepen understanding and enhance learning, aiding in the creation of interconnected curriculums.

\textbf{Task 6: Idea Hamster}
The task prompts participants to develop project proposals by applying learned concepts to real-world contexts, providing examples and outcomes to fuel creativity.


\subsection{KG-Enhanced Model}
To address TutorQA tasks, we first utilize Graphusion framework to construct an NLP KG. Then we design an enhanced framework for the interaction between the LLM and the concept graph, which includes two steps: command query and answer generation. In the command query stage, an LLM independently generates commands to query the concept graph upon receiving the query, thereby retrieving relevant paths. During the answer generation phase, these paths are provided to the LLM as contextual prompt, enabling it to perform concept graph reasoning and generate answers.

\subsection{Evaluation}
\textbf{Accuracy} We report accuracy score for Task 1 and Task 4, as they are binary classification tasks.

\textbf{Similarity score} 
For Tasks 2 and 3, the references consist of a list of concepts. Generally, LLMs demonstrate creativity by answering with novel concepts, which are often composed of more contemporary and fresh words, even though they might not exactly match the words in the concept graph. Consequently, conventional evaluation metrics like keyword matching are unsuitable for these tasks. To address this, we propose the \textbf{similarity score}. This metric calculates the semantic similarity between the concepts in the predicted list $C_{pred}$ and the ground truth list $C_{gold}$. Specifically, as shown in Eq.~\ref{eq:sim}, for a concept $m$ from the predicted list, and a concept $n$ from the ground truth list, we calculate the cosine similarity between their embeddings achieved from pre-trained BERT model ~\cite{Devlin2019BERTPO}. We then average these similarity scores to obtain the similarity score. 

\begin{equation}
\begin{aligned}
& Score= \frac{\sum_{m \in C_{pred}} \sum_{n \in C_{gold}} sim(m,n)}{|C_{pred}| \times |C_{gold}|}.
\nonumber
\label{eq:sim}
\end{aligned}
\end{equation}
\vspace{3mm}

By averaging the similarity scores, the final score provides a comprehensive measure of the overall semantic alignment between the predicted and ground truth concepts.

\textbf{Hit Rate} For Task 5, we employ the classical Hit Rate metric, expressed as a percentage. This measure exemplifies the efficiency of LLM at retrieving and presenting relevant concepts in its output as compared to a provided list of target concepts.


Our comparative analysis in Table~\ref{tab:tutorqa_res}, which features GPT-4o as the base model, reveals significant improvement across Tasks 1 to 5. These findings underscore the pivotal role of integrating a KG into our pipeline, solidly confirming our assertion that knowledge-augmented systems exhibit exceptional QA capabilities. The marked enhancements are indicative of the substantial potential that KGs hold for advancing NLP applications, especially in educational contexts where such systems can tailor and enhance the learning experience with deeper, context-relevant insights.

\begin{table}[ht]
\centering
\small
\begin{tabular}{lccccc}
\toprule
\textbf{Setting}     & \textbf{T1} &\textbf{ T2 }& \textbf{T3} & \textbf{T4}& \textbf{T5} \\
\midrule
Zero-shot  & 69.20 & 64.42  & 66.61 & 44.00  & 11.45  \\   
Ours  & \textbf{92.00} & \textbf{80.29} & \textbf{77.85} & \textbf{50.00}  & \textbf{15.65} \\ 
\bottomrule
\end{tabular}
\caption{Evaluation of TutorQA on Tasks 1-5. T1, T4: accuracy; T2, T3: similarity score; T5: hit rate.}
\label{tab:tutorqa_res}
\vspace{-4mm}
\end{table}


\paragraph{Expert Evaluation} In Task 6, where open-ended answers are generated without gold-standard responses, we resort to expert evaluation for comparative analysis between baseline results and our model. Despite available LLM-centric metrics like G-Eval~\cite{Liu2023GEvalNE}, the specific evaluation needs of this task warrant distinct criteria, particularly examining the persuasive and scientifically sound elements of generated project proposals. Four evaluation criteria, rated on a 1-5 scale, are employed: \textit{Concept Relevancy}: the project's alignment with the query concepts. \textit{Concept Coverage}: the extent to which the project encompasses the query concepts. \textit{Project Convincity}: the persuasiveness and practical feasibility of the project. \textit{Scientific Factuality}: the scientific accuracy of the information within the project. Evaluation by two NLP experts, with a Kappa score of 0.6689, suggests substantial agreement. The results in Table~\ref{tab:tutorqa_res_6} indicate that while both settings achieve high scores across all criteria, our pipeline exhibits a marginally superior performance, particularly in terms of Convincity and Factuality. This suggests that our pipeline might be better at generating content that is not only factually accurate but also presents it in a way that is more persuasive to the reader.

\begin{table}[ht]
\centering
\small
\setlength{\tabcolsep}{2pt}
\begin{tabular}{@{}lcccc@{}}

\toprule
\textbf{ Model}     & \textbf{Relevancy} &\textbf{ Coverage }& \textbf{Convincity} & \textbf{Factuality} \\
\midrule
Zero-shot  & 4.750 & 4.840 & 4.380 & 4.625 \\
Ours & \textbf{4.845} & \textbf{4.905} & \textbf{4.720} & \textbf{4.770}  \\
\bottomrule
\end{tabular}
\caption{Expert evaluation of TutorQA on Task 6.}
\label{tab:tutorqa_res_6}
\vspace{-4mm}
\end{table}




\section{Ablation Study and Analysis}

\subsection{RAG Data for Link Prediction} 
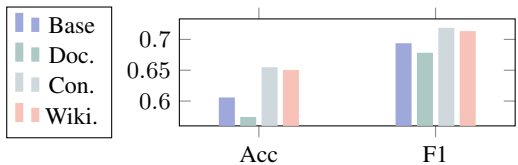
\begin{figure}[ht]
    \centering
    
\begin{tikzpicture}
\centering
\small
\begin{axis}[
    ybar,
    bar width=6pt,
    enlarge x limits=0.45,
    legend style={
        at={(-0.2,0.5)}, 
        anchor=east,     
        legend columns=1 
    },
    symbolic x coords={Acc, F1},
    xtick=data,
    height=3cm,
    width=6cm,
]

\addplot[fill=color1, draw opacity=0, draw=none] coordinates {(Acc,0.6058) (F1,0.6937)};
\addlegendentry{Base}

\addplot[fill=color2, draw=none, draw opacity=0] coordinates {(Acc,0.5742) (F1,0.6783) };
\addlegendentry{Doc.}

\addplot[fill=color3, draw=none, draw opacity=0] coordinates {(Acc,0.6548) (F1,0.7187) };
\addlegendentry{Con.}

\addplot[fill=color4, draw=none, draw opacity=0] coordinates {(Acc,0.6503) (F1,0.7133) };
\addlegendentry{Wiki.}
\end{axis}
\end{tikzpicture}
\vspace{-2mm}
\caption{Link Prediction Ablation Study: Comparison of models with external data.}
\label{fig:info_comp}
\vspace{-4mm}
\end{figure}

We explore the potential of external data in enhancing concept graph recovery. This is achieved by expanding the \texttt{\{Additional Information\}} part in the \textbf{LP Prompt}. We utilize LLaMa as the \textbf{Base} model, focusing on the NLP domain. We introduce three distinct settings: \textbf{Doc.}: In-domain lecture slides data as free-text; \textbf{Con.}: Adding one-hop neighboring concepts from the training set as additional information related to the query concepts. \textbf{Wiki.}: Incorporating the introductory paragraph of the Wikipedia page of each query concept. As illustrated in Fig \ref{fig:info_comp}, our findings indicate that incorporating LectureBankCD documents (Doc.) significantly diminishes performance. This decline can be attributed to the introduction of noise and excessively lengthy content, which proves challenging for the LLM to process effectively. Conversely, the inclusion of neighboring concepts (Con.) markedly enhances the base model's performance. However, it relies on training data, rendering it incompatible with our primary focus on the zero-shot setting. Incorporating Wikipedia content also yields improvements and outperforms the use of LectureBankCD, likely due to higher text quality. 

\subsection{Graphusion Modules} 
\begin{figure}[ht]
\begin{center}
\begin{tikzpicture}
\small
\begin{axis}[
        ybar=-0.5cm, 
        bar width=0.6cm, 
        enlarge x limits=0.2, 
        ylabel={Rating},
        symbolic x coords={LP,LPRAG,Extraction,Graphusion},
        xtick={LP,LPRAG,Extraction,Graphusion}, 
        ymin=0,
        ymax=3.5,
        height=3cm,
        width=8cm,
        yticklabels={,,}, 
        legend style={draw=none}, 
        xticklabel style={anchor=north, rotate=0, font=\footnotesize}, 
        nodes near coords,
    ]
    \addplot[fill=color1, draw=none] coordinates {(LP,2.07)};
    \addplot[fill=color2, draw=none] coordinates {(LPRAG,2.12)};
    \addplot[fill=color3, draw=none] coordinates {(Extraction,1.80)};
    \addplot[fill=color4, draw=none] coordinates {(Graphusion, 2.37)};
\end{axis}
\end{tikzpicture}
\vspace{-2mm}
    \caption{Ablation study on Graphusion modules: We compare four settings with GPT-4o as base. }
    \label{fig:ablation_gf}
    \end{center}
\vspace{-4mm}
\end{figure}
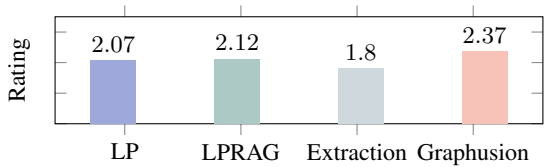

We conduct an ablation study on the KGC task by comparing different settings, as shown in Fig ~\ref{fig:ablation_gf}. We evaluate four configurations: Link Prediction using the LP prompt (\textit{LP}), Link Prediction with RAG (\textit{LPRAG}), Candidate Triplet Extraction without Fusion (\textit{Extraction}), and Graphusion (\textit{Graphusion}). In the first two settings, we implement a straightforward scenario where concept pairs are provided, and the relationship is predicted directly through link prediction. All experiments are conducted using GPT-4 as the base language model. We report the average human evaluation rating on relation quality. The concept entities remain fixed, so their ratings are not included. Our findings indicate that the Graphusion framework achieves the best performance. Removing the core fusion component (the \textit{Extraction} setting) significantly diminishes performance, underscoring the effectiveness of the fusion module.

\subsection{Graphusion Case Study}
\begin{figure}[ht]
    \centering
    \includegraphics[width=0.5\textwidth]{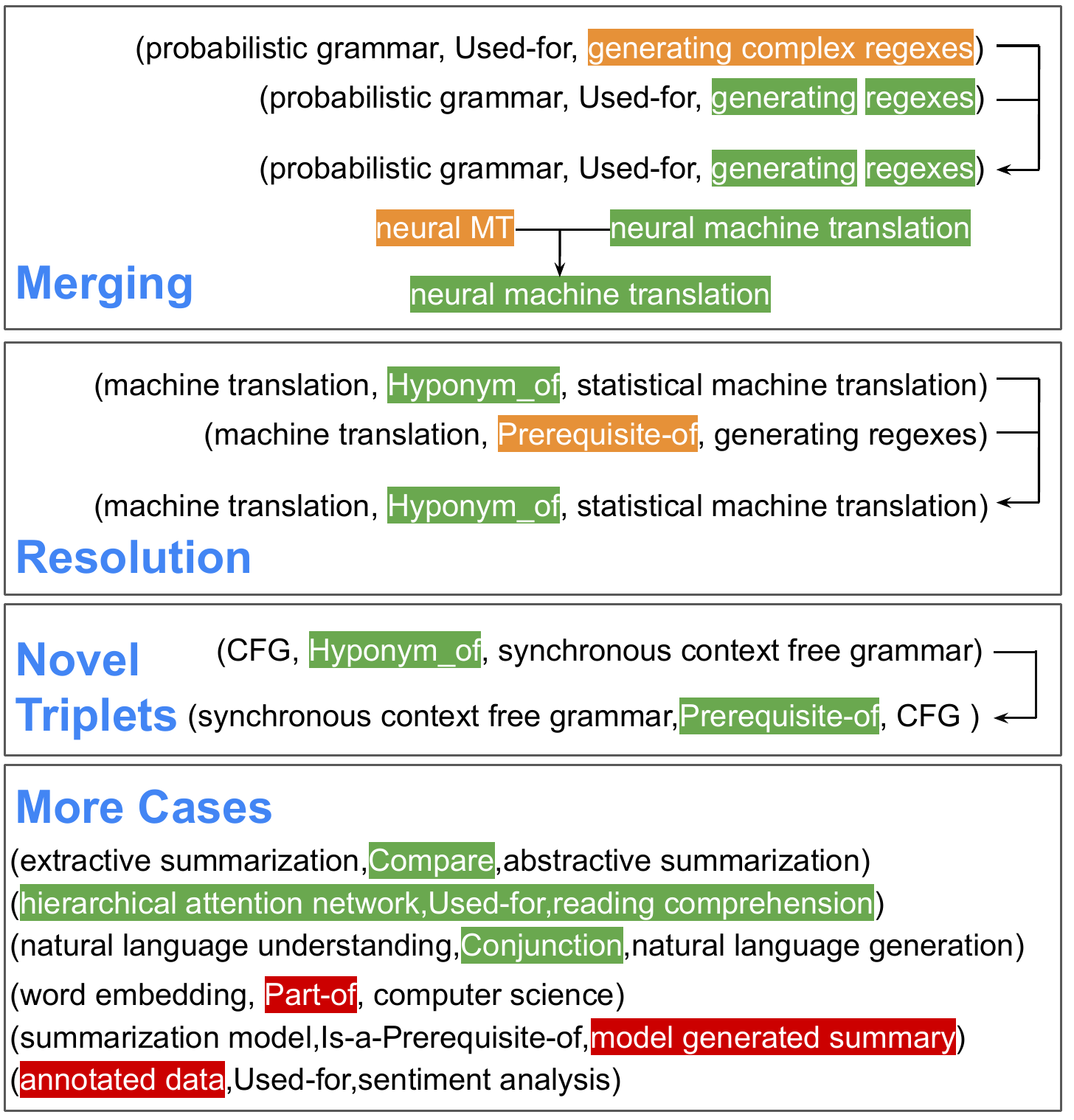}
    \caption{Case studies for Graphusion on the GPT-4o model: Correct parts are highlighted in green, resolved and merged parts in orange, and incorrect parts in red. }
    \label{fig:graphusion_case}
    \vspace{-6mm}
\end{figure}

In Fig~\ref{fig:graphusion_case}, we present case studies from our Graphusion framework using GPT-4o. Graphusion effectively merges similar concepts (\texttt{neural MT} and\texttt{ neural machine translation}) and resolves relational conflicts (\texttt{prerequisite of} vs \texttt{hyponym of}). Additionally, it can infer novel triplets absent from the input. We highlight both positive and negative outputs from Graphusion. For instance, it correctly identifies that a technique is used for a task (\texttt{hierarchical attention network, used for, reading comprehension}). However, it may make mistakes in concept recognition, such as concepts with poor granularity (\texttt{annotated data}, \texttt{model generated summary}) and identifying incorrect relations (\texttt{word embedding} being inaccurately categorized as part of \texttt{computer science}).

\subsection{TutorQA Task 2 \& Task 3: Concept Entity Counts} As depicted in Fig ~\ref{fig:task2,3}, We evaluate the average number of concept entities generated by GPT-4o and our Graphusion framework in the responses for Task 2 and Task 3. 
The results show that without the enhancement of KG, GPT-4o tends to generate more concept entities (Task 2: 11.04, Task 3: 11.54), many of which are irrelevant or broad. In contrast, our Graphusion framework generates more accurate and targeted concept entities. 

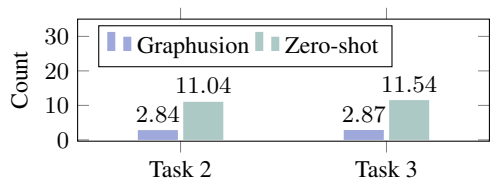
\begin{figure}[ht]
    \centering
        \begin{tikzpicture}
        \small
        \begin{axis}[
            ybar,
            bar width=15pt, 
            enlarge x limits=0.5, 
            legend style={
                at={(0.05,0.95)}, 
                anchor=north west,
                legend columns=2,
            },
            ylabel={Count},
            ymax = 35,
            symbolic x coords={Task 2, Task 3},
            xtick=data,
            height=3.2cm, 
            width = 7cm,
            nodes near coords, 
            nodes near coords align={vertical},
        ]
        \addplot[fill=color1, draw=none, draw opacity=0] coordinates {(Task 2, 2.84) (Task 3,2.87)};
        \addlegendentry{Graphusion}
        \addplot[fill=color2, draw=none, draw opacity=0] coordinates {(Task 2, 11.04) (Task 3, 11.54)};
        \addlegendentry{Zero-shot}
        \end{axis}
            \vspace{-2mm}
        \end{tikzpicture}
    \caption{Comparison of concept counts in Task 2 and Task 3.}
    \label{fig:task2,3}
    \vspace{-4mm}
\end{figure}

\section{Conclusion}

In this work, we explored the application of LLMs for scientific KG fusion and construction. Initially, we developed Graphusion, which enables LLMs to perform zero-shot KGC from free text. Subsequently, we introduced TutorQA, an expert-verified, NLP-centric benchmark designed for QA using a concept graph. Lastly, we devised a pipeline aimed at augmenting QA performance by leveraging LLMs and constructed KG.



\section*{Limitations}
\textbf{Graph Construction} 
Constructing a KG from free-text, especially under a zero-shot context, relies on the quality and scale of the corpus. In this paper, we showcased that applying paper abstracts is a possible way. While we did not have a chance to test other data such as text books or web posts. Besides, evaluation is challenging as it is hard to construct a standard test set. So our evaluation was mostly conducted by human experts with a reasonable scale. 

\textbf{Evaluation metrics of TutorQA} For Task 2 and 3, LLMs often generate novel concepts in their responses. To address this, we evaluated answers based on semantic similarities to compute a score. A notable limitation is the disregard for concept order in the provided answer paths. Addressing this concern will be a focus of our future work.

\section*{Ethical Considerations}
In our research, we have meticulously addressed ethical considerations, particularly regarding our dataset TutorQA and Graphusion framework. TutorQA has been expert-verified to ensure it contains no harmful or private information about individuals, thereby upholding data integrity and privacy standards. Our methods, developed on publicly available Large Language Models optimized for text generation, adhere to established ethical norms in AI research. We recognize the potential biases in such models and are committed to ongoing monitoring to prevent any unethical content generation, thereby maintaining the highest standards of research integrity and responsibility.

\bibliography{enmlp_main}
\appendix
\onecolumn

\section{Prompt Templates}
\label{app:prompt}

\subsection{Main Framework}
\label{app:main_prompt}
\vspace{5mm}
\paragraph{LP Prompt}  
\mbox{}
\begin{lstlisting}[language={}, captionpos=b, label={lst: lp_1}, basicstyle=\linespread{0.85}\ttfamily, frame=single, breaklines=true, keepspaces=true]
We have two {domain} related concepts: A: {concept_1} and B: {concept_2}.

Do you think learning {concept_1} will help in understanding {concept_2}?

Hints:
1. Answer YES or NO only.
2. This is a directional relation, which means if the answer is "YES", (B, A) is 
   false, but (A, B) is true.
3. Your answer will be used to create a knowledge graph.
   
{Additional Information}
\end{lstlisting}
\vspace{5mm}

\paragraph{LP Prompt With Chain-of-Thought}
\mbox{}
\begin{lstlisting}[language={}, captionpos=b, label={lst: lp_cot}, basicstyle=\linespread{0.85}\ttfamily, frame=single, breaklines=true, keepspaces=true]
We have two {domain} related concepts: A: {concept_1} and B: {concept_2}. 

Assess if learning {concept_1} is a prerequisite for understanding {concept_2}. 

Employ the Chain of Thought to detail your reasoning before giving a final answer.

# Identify the Domain and Concepts: Clearly define A and B within their domain. 
  Understand the specific content and scope of each concept.
  
# Analyze the Directional Relationship: Determine if knowledge of concept A is 
  essential before one can fully grasp concept B. This involves considering if A 
  provides foundational knowledge or skills required for understanding B.
  
# Evaluate Dependency: Assess whether B is dependent on A in such a way that 
  without understanding A, one cannot understand B.

# Draw a Conclusion: Based on your analysis, decide if understanding A is a 
  necessary prerequisite for understanding B.
  
# Provide a Clear Answer: After detailed reasoning, conclude with a distinct answer
  : <result>YES</result> if understanding A is a prerequisite for understanding B, 
  or <result>NO</result> if it is not.   
\end{lstlisting}

\clearpage
\paragraph{Extraction Prompt}
\mbox{}
\begin{lstlisting}[language={}, captionpos=b, label={lst: ep}, basicstyle=\linespread{0.85}\ttfamily, frame=single, breaklines=true, keepspaces=true]
### Instruction:
You are a domain expert in natural language processing, and now you are building a 
knowledge graph in this domain. 

Given a context (### Content), and a query concept (### Concept), do the following:

1. Extract the query concept and in-domain concepts from the context, which should 
   be fine-grained: could be introduced by a lecture slide page, or a whole 
   lecture, or possibly to have a Wikipedia page. 

2. Determine the relations between the query concept and the extracted concepts, in 
   a triplet format: (<head concept>, <relation>, <tail concept>). The relation 
   should be functional, aiding learners in understanding the knowledge. The query 
   concept can be the head concept or tail concept. 

   We define 7 types of the relations:

   a) Compare: Represents a relation between two or more entities where a 
      comparison is being made. For example, "A is larger than B" or "X is more 
      efficient than Y."
    
   b) Part-of: Denotes a relation where one entity is a constituent or component of 
      another. For instance, "Wheel is a part of a Car."
    
   c) Conjunction: Indicates a logical or semantic relation where two or more 
      entities are connected to form a group or composite idea. For example, "Salt 
      and Pepper."
    
   d) Evaluate-for: Represents an evaluative relation where one entity is assessed 
      in the context of another. For example, "A tool is evaluated for its 
      effectiveness."
    
   e) Is-a-Prerequisite-of: This dual-purpose relation implies that one entity is 
      either a characteristic of another or a required precursor for another. For 
      instance, "The ability to code is a prerequisite of software development."
    
   f) Used-for: Denotes a functional relation where one entity is utilized in 
      accomplishing or facilitating the other. For example, "A hammer is used for 
      driving nails."
    
   g) Hyponym-Of: Establishes a hierarchical relation where one entity is a more 
      specific version or subtype of another. For instance, "A Sedan is a hyponym 
      of a Car."

3. Please note some relations are strictly directional. For example, "A tool is 
   evaluated for B" indicates (A, Evaluate-for, B), NOT (B, Evaluate-for, A). 
   Among the seven relation types, only "a) Compare" and "c) Conjunction" are not 
   direction-sensitive.

4. You can also extract triplets from the extracted concepts, and the query concept 
   may not be necessary in the triplets. 
   
5. Your answer should ONLY contain a list of triplets, each triplet is in this 
   format: (concept, relation, concept). For example: "(concept, relation, concept)
   (concept, relation, concept)." No numbering and other explanations are needed. 
   
6. If ### Content is empty, output None. 
\end{lstlisting}

\clearpage
\paragraph{Fusion Prompt}
\mbox{}
\begin{lstlisting}[language={}, captionpos=b, label={lst: fp}, basicstyle=\linespread{0.85}\ttfamily, frame=single, breaklines=true, keepspaces=true]
### Instruction: You are a knowledge graph builder. 
    Now please fuse two sub-knowledge graphs about the concept "{concept}".

Graph 1: {LLM-KG}   Graph 2: {E-G}

Rules for Fusing the Graphs:
1. Union the concepts and edges. 

2. If two concepts are similar, or refer to the same concept, merge them into one 
   concept, keeping he one that is meaningful or specific. For example, "lstm" 
   versus "long short-term memory",  please keep "long short-term memory". 
   
3. Only one relation is allowed between two concepts. If there is a conflict, read 
   the "### Background" to help you keep the correct relation. knowledge to keep the 
   correct one. For example, (ROUGE, Evaluate-for, question answering model) and 
   (ROUGE,Used-for , question answering model) are considered to be conflicts. 

4. Once step 3 is done, consider every possible concept pair not covered in step 2. 
   For example, take a concept from Graph 1, and match it from Graph 2. Then, 
   please refer to "### Background" to summarize new triplets.   

Hint: the relation types and their definition. You can use it to do Step 3.
We define 7 types of the relations:

   a) Compare: Represents a relation between two or more entities where a 
      comparison is being made. For example, "A is larger than B" or "X is more 
      efficient than Y."
    
   b) Part-of: Denotes a relation where one entity is a constituent or component of 
      another. For instance, "Wheel is a part of a Car."
    
   c) Conjunction: Indicates a logical or semantic relation where two or more 
      entities are connected to form a group or composite idea. For example, "Salt 
      and Pepper."
    
   d) Evaluate-for: Represents an evaluative relation where one entity is assessed 
      in the context of another. For example, "A tool is evaluated for its 
      effectiveness."
    
   e) Is-a-Prerequisite-of: This dual-purpose relation implies that one entity is 
      either a characteristic of another or a required precursor for another. For 
      instance, "The ability to code is a prerequisite of software development."
    
   f) Used-for: Denotes a functional relation where one entity is utilized in 
      accomplishing or facilitating the other. For example, "A hammer is used for 
      driving nails."
    
   g) Hyponym-Of: Establishes a hierarchical relation where one entity is a more 
      specific version or subtype of another. For instance, "A Sedan is a hyponym 
      of a Car."
   
### Background: 
{background}

### Output Instruction: 
    Output the new merged data by listing the triplets. Your answer should ONLY contain triplets in this format: (concept, relation, concept). No other explanations or numbering are needed. Only triplets, no intermediate results.
\end{lstlisting}



\subsection{Ablation Study}
\label{app:ext_prompt}
\vspace{5mm}

\paragraph{Link Prediction with \textbf{Doc.} } 
\mbox{}
\begin{lstlisting}[language={}, captionpos=b, label={lst: lpwd}, basicstyle=\linespread{0.85}\ttfamily, frame=single, breaklines=true, keepspaces=true]
We have two {domain} related concepts: A: {concept_1} and B: {concept_2}.

Do you think learning {concept_1} will help in understanding {concept_2}?

Hints:
1. Answer YES or NO only.
2. This is a directional relation, which means if the answer is "YES", (B, A) is 
   false, but (A, B) is true.
3. Your answer will be used to create a knowledge graph.
   
And here are related contents to help: 
{related documents concatenation}
\end{lstlisting}

\paragraph{Link Prediction with \textbf{Con.}} 
\mbox{}
\begin{lstlisting}[language={}, captionpos=b, label={lst: lpwc}, basicstyle=\linespread{0.85}\ttfamily, frame=single, breaklines=true, keepspaces=true]
We have two {domain} related concepts: A: {concept_1} and B: {concept_2}.

Do you think learning {concept_1} will help in understanding {concept_2}?

Hints:
1. Answer YES or NO only.
2. This is a directional relation, which means if the answer is "YES", (B, A) is 
   false, but (A, B) is true.
3. Your answer will be used to create a knowledge graph.
   
And here are related contents to help: 

We know that {concept_1} is a prerequisite of the following concepts: 
{1-hop successors of concept_1 from training data};

The following concepts are the prerequisites of {concept_1}:
{1-hop predecessors of concept_1 from training data}. 

We know that {concept_2} is a prerequisite of the following concepts: 
{1-hop successors of concept_2 from training data};

The following concepts are the prerequisites of {concept_2}:
{1-hop predecessors of concept_2 from training data}. 
\end{lstlisting}

\paragraph{Link Prediction with \textbf{Wiki.}} 

\mbox{}
\begin{lstlisting}[language={}, captionpos=b, label={lst: lpww}, basicstyle=\linespread{0.85}\ttfamily, frame=single, breaklines=true, keepspaces=true]
We have two {domain} related concepts: A: {concept_1} and B: {concept_2}.

Do you think learning {concept_1} will help in understanding {concept_2}?

Hints:
1. Answer YES or NO only.
2. This is a directional relation, which means if the answer is "YES", (B, A) is 
   false, but (A, B) is true.
3. Your answer will be used to create a knowledge graph.
   
And here are related contents to help: 
{Wikipedia introductory paragraph of {concept_1}}
{Wikipedia introductory paragraph of {concept_2}}
\end{lstlisting}


\section{Experimental Setup}
\subsection{Experiments} In our experimental setup, we employed Hugging Face's \texttt{LLaMA-2-70b-chat-hf}\footnote{\url{https://huggingface.co/meta-LLaMA}} and 
\texttt{LLaMA-3-70b-chat-hf}\footnote{\url{https://huggingface.co/meta-LLaMA/Meta-LLaMA-3-70B}} for LLaMA2 and LLaMA3 on a cluster equipped with 8 NVIDIA A100 GPUs. For GPT-3.5 and GPT-4, we used OpenAI's \texttt{gpt-3.5-turbo} \texttt{gpt-4-1106-preview}, and \texttt{gpt-4o} APIs, respectively, each configured with a temperature setting of zero. The RAG models are implemented using Embedchain~\cite{embedchain}. To solve TutorQA tasks, we implemented our pipeline using LangChain\footnote{\url{https://www.langchain.com/}}. The total budget spent on this project, including the cost of the GPT API service, is approximately 500 USD.


\subsection{Additional Corpora Description}
\textbf{TutorialBank} We obtained the most recent version of TutorialBank from the authors, which consists of 15,583 manually curated resources. This collection includes papers, blog posts, textbook chapters, and other online resources. Each resource is accompanied by metadata and a publicly accessible URL. We downloaded the resources from these URLs and performed free text extraction. Given the varied data formats such as PDF, PPTX, and HTML, we encountered some challenges during text extraction. To ensure text quality, we filtered out sentences shorter than 25 words. Ultimately, this process yielded 559,217 sentences suitable for RAG and finetuning experiments.\\
\noindent\textbf{NLP-Papers} We downloaded conference papers from EMNLP, ACL, and NAACL spanning the years 2021 to 2023. Following this, we utilized Grobid (\url{https://github.com/kermitt2/grobid}) for text extraction, resulting in a collection of 4,787 documents with clean text.

\clearpage
\section{Link Prediction}

\subsection{Experiments}
Since LectureBankCD contains data from three domains: NLP, computer vision (CV), and bioinformatics (BIO), we further compare the performance across all the domains, presenting the results in Tab.~\ref{tab:main}. Specifically, the RAG data predominantly consists of NLP-related content, which is why there is no noticeable improvement in the CV and BIO domains when using RAG.

\begin{table*}[h]
\small
\centering
\begin{tabular}{lcccccccc}
\toprule
\textbf{Method} & \multicolumn{2}{c}{\textbf{NLP}} & \multicolumn{2}{c}{\textbf{CV}} & \multicolumn{2}{c}{\textbf{BIO}} & \multicolumn{2}{c}{\textbf{Overall}} \\
\cmidrule(lr){2-3} \cmidrule(lr){4-5} \cmidrule(lr){6-7} \cmidrule(lr){8-9}
 & \textbf{Acc} & \textbf{F1} & \textbf{Acc} & \textbf{F1} & \textbf{Acc} & \textbf{F1} & \textbf{Acc} & \textbf{F1} \\
\midrule 
\multicolumn{9}{l}{\textit{Supervised Baselines}} \\
P2V & 0.6369 & 0.5961 & 0.7642 & 0.7570 & 0.7200 & 0.7367 & 0.7070 & 0.6966 \\
BERT & 0.7088 & 0.6963 & 0.7572 & 0.7495 & 0.7067 & 0.7189 & 0.7242 & 0.7216 \\
DeepWalk & 0.6292 & 0.5860 & 0.7988 & 0.7910 & 0.7911 & 0.8079 & 0.7397 & 0.7283 \\
Node2vec & 0.6209 & 0.6181 & 0.8197 & 0.8172 & 0.7956 & 0.8060 & 0.7454 & 0.7471 \\
\midrule[0.2pt]
\multicolumn{9}{l}{\textit{LLM - Supervised}} \\
LLaMA - Classifier & 0.6684 & 0.6475 & 0.6184 & 0.6549 & 0.6043 & 0.6644 & 0.6304 & 0.6556 \\
LLaMA - Graph & \underline{0.7174} & \underline{0.7673} & \underline{0.7839} & \underline{0.8223} & \underline{0.8217} & \underline{0.8464} & \underline{0.7743} & \underline{0.8120}\\
\midrule
\multicolumn{9}{l}{\textit{Zero-shot (zs)}} \\
LLaMA & 0.6058 & 0.6937 & 0.6092 & 0.6989 & 0.6261 & 0.6957 & 0.6137 & 0.6961 \\
GPT-3.5 & 0.6123 & 0.7139 & 0.6667 & 0.7271 & 0.6696 & 0.6801 & 0.6495 & 0.7070 \\
GPT-4 & 0.7639 & 0.7946 & \textbf{0.7391 }& \textbf{0.7629} & \textbf{0.7348} & \textbf{0.7737} & \textbf{0.7459} & \textbf{0.7771} \\
\midrule                            
\multicolumn{9}{l}{\textit{Zero-shot + RAG}} \\
GPT-3.5 & 0.7587 & 0.7793 & 0.6828 & 0.7123 & 0.6870 & 0.7006 & 0.7095 & 0.7307 \\
GPT-4 & \textbf{0.7755} & \textbf{0.7958} & 0.7230 & 0.7441 & 0.7174 & 0.7200 & 0.7386 & 0.7533 \\
\bottomrule
\end{tabular}
\caption{Link prediction results across all domains on the LectureBankCD test set: We present accuracy (Acc) and F1 scores. Bolded figures indicate the best performance in the zero-shot setting, while underlined scores represent the highest achievements in the supervised setting. In this paper, we apply LLaMA2-70b for all experiments.}
\label{tab:main}
\vspace{-3mm}
\end{table*}

\subsection{Ablation Study}

\textbf{Prompting Strategies} In Tab. ~\ref{tab: cot}, we explore the impact of different prompting strategies for concept graph recovery, comparing CoT and zero-shot prompts across both NLP and CV domains. The results indicate the introduction of CoT is not improving. We further find that CoT Prompting more frequently results in negative predictions. This finding serves as a drawback for our study, as it somewhat suppresses the performance of our system. This observation highlights the need to balance the impact of CoT on the rigor and complexity of predictions, especially in the context of graph recovery.
\begin{table}[h]
\centering
\small
\begin{tabular}{lcccc}
\toprule
  Model          & \multicolumn{2}{c}{NLP}  &  \multicolumn{2}{c}{CV} \\ 
\cmidrule(r){2-3} \cmidrule(lr){4-5} 
     & Acc & F1  & Acc & F1  \\
\midrule
GPT-4 zs     & 0.7639 & 0.7946 & 0.7391&	0.7629 \\
GPT-4 CoT  & 0.7342 & 0.6537 &  0.6122 &0.4159 \\
\bottomrule
\end{tabular}
\caption{Comparison of zero-shot and CoT prompts with GPT-4: Results on NLP and CV. }
\label{tab: cot}
\end{table}
\clearpage

\noindent\textbf{Finetuning} We further explore the impact of finetuning on additional datasets, with results detailed in Table~\ref{tab:finetuning}. Specifically, we utilize LLaMA2-70b~\cite{touvron2023llama}, finetuning it on two previously mentioned datasets: TutorialBank and NLP-Papers. Both the zero-shot LLaMA and the finetuned models are employed to generate answers. As these answers are binary (\texttt{YES} or \texttt{NO}), we can calculate both the accuracy and F1 score for evaluation. However, the results indicate that finetuning does not yield positive outcomes. This can be attributed to two potential factors: 1) the poor quality of data, and 2) limited effectiveness in aiding the graph recovery task. We leave this part as the future work. 

\begin{table*}[h]
\small
\centering
\begin{tabular}{lcc}
\toprule
\textbf{Dataset} & \textbf{Acc} & \textbf{F1} \\
\midrule
LLaMA2-70b & \textbf{0.6058} & \textbf{0.6937} \\	
TutorialBank & 0.4739  & 0.0764 \\
NLP Papers & 0.5435 & 0.6363 \\
\bottomrule
\end{tabular}
\caption{Comparison of the effect of finetuning: Results on NLP domain.}
\label{tab:finetuning}
\end{table*}

\section{Graphusion: human evaluation rubric}
\begin{enumerate}
    \item Concept Entity Quality
    \begin{itemize}
        \item \textbf{Excellent (3 points)}: Both concepts are highly relevant and specific to the domain. At an appropriate level of detail, neither too broad nor too specific. For example, a concept could be introduced by a lecture slide page, or a whole lecture, or possibly have a Wikipedia page.
        \item \textbf{Acceptable (2 points)}: Concept is somewhat relevant, or granularity is acceptable.
        \item \textbf{Poor (1 point)}: Concept is at an inappropriate level of detail, too broad or too specific.
    \end{itemize}
    \item Relation Quality
    \begin{itemize}
        \item \textbf{Correct (3 points)}: The relation logically and accurately describes the relationship between the head and tail concepts.
        \item \textbf{Somewhat Correct (2 points)}: The relation is acceptable but has minor inaccuracies or there might be another better or correct answer.
        \item \textbf{Incorrect (1 point)}: The relation does not logically describe the relationship between the concepts.
    \end{itemize}
\end{enumerate}

\clearpage
\section{TutorQA}
\label{app:tutorqa_info}

\subsection{Benchmark Details}
We show the data analysis in Tab.~\ref{tab:data_stats}. 

\begin{table}[h]
\small
\centering
\begin{tabular}{cccccccc}
\toprule
\textbf{Task} & \multicolumn{3}{c}{\textbf{Question Token}} & \multicolumn{3}{c}{\textbf{Concept Count}}  & \textbf{Number}\\ 
\cmidrule(lr){2-4} \cmidrule(lr){5-7} 
 & \textbf{Max} & \textbf{Min} & \textbf{Mean} & \textbf{Max} & \textbf{Min} & \textbf{Mean} &  \\ \midrule
T1 & 77 & 61 & 68.00 & - & - & - & 250\\ 
T2 & 27 & 22 & 23.48 & 7 & 1 & 1.79  & 250\\ 
T3 & 40 & 34 & 36.66 & 8 & 2 & 3.36 & 250\\ 
T4 & 88 & 76 & 83.00 & - & - & - & 250\\ 
T5 & 21 & 18 & 19.26 & 8 & 1 & 4.76 & 100\\ 
T6 & 54 & 42 & 48.62 & - & - & - & 100\\ \bottomrule
\end{tabular}
\caption{TutorQA data statistics comparison: The answers in T1 are only "True" or "False", and the answers in T4 are relations, while the answers in T6 are free text with open-ended answers.}
\label{tab:data_stats}
\end{table}

\subsection{Task 2 and Task 3: case study}

\begin{table*}[h]
\centering
\small
\begin{tabularx}{\textwidth}{lX}
\toprule
\textbf{\textit{Question} [Task2]} & \textit{In the field of Natural Language Processing, I want to learn about \textbf{multilingual model}. What concepts should I learn first?} \\
\midrule
\textbf{GPT-4o} & \textbf{Tokenization, Embeddings, Transfer Learning, Cross-lingual Transfer, Zero-shot Learning, Multilingual Corpora, Language Modeling, Fine-tuning, Evaluation Metrics, Pretrained Models} \\
\textbf{Ours} & \textbf{language models, machine translation, cross-lingual embeddings, transfer learning, tokenization, fine-tuning} \\
\midrule
\textbf{\textit{Question} [Task3]} & \textit{In the field of Natural Language Processing, I know about \textbf{natural language processing intro}, now I want to learn about \textbf{t-sne}. What concept path should I follow?}\\
\midrule
\textbf{GPT-4o} & \textbf{natural language processing, dimensionality reduction, t-SNE, perplexity, high-dimensional data, data visualization, machine learning} \\
\textbf{Ours} & \textbf{natural language processing intro, vector representations, t-sne} \\
\bottomrule
\end{tabularx}
\caption{Case study on TutorQA Task 2 and Task 3: GPT-4o, and GPT-4o-Graphusion.}
\label{tab:case_study_task2,3}
\vspace{-2mm}
\end{table*}

\subsection{Task 4: case study}
\begin{table*}[h]
\centering
\small
\begin{tabularx}{\textwidth}{lX}
\toprule
\textbf{\textit{Question}} & \textit{Given the following edges constituting a concept subgraph, please identify and select the possible type of relationship between \textbf{natural language generation} and \textbf{natural language understanding}.} \\

\midrule
\textbf{GPT-4o} & \textbf{Is-a-Prerequisite-of} \\
\textbf{Ours} & \textbf{Conjunction} \\

\bottomrule
\end{tabularx}
\caption{Case study on TutorQA Task 4: GPT-4o, and GPT-4o-Graphusion.}
\label{tab:case_study_task4}
\vspace{-2mm}
\end{table*}

\subsection{Task 6: Human Evaluation Rubrics} 
\label{app:case}

\begin{itemize}
    \item \textbf{Concept Relevancy (1-5):}
    \begin{itemize}
        \item Rate how well the project description aligns with the provided query concepts.
        \item Score 1 if the project is not related to any of the query concepts.
        \item Score 5 if the project directly addresses and is entirely based on the provided query concepts.
    \end{itemize}

    \item \textbf{Concept Coverage (1-5):}
    \begin{itemize}
        \item Evaluate how comprehensively the project description covers the query concepts.
        \item Score 1 if only a minimal or superficial coverage of the concepts is present.
        \item Score 5 if all query concepts are thoroughly and extensively discussed in the project description.
    \end{itemize}

    \item \textbf{Project Convincity (1-5):}
    \begin{itemize}
        \item Assess the overall persuasiveness and viability of the project described.
        \item Score 1 if the project seems implausible, lacks a clear purpose, or doesn't appear convincing.
        \item Score 5 if the project is highly convincing, well-structured, and appears to have long-term durability.
    \end{itemize}

    \item \textbf{Scientific Factuality (1-5):}
    \begin{itemize}
        \item Determine the accuracy and correctness of the scientific details presented in the project description.
        \item Score 1 if the content contains multiple factual inaccuracies or appears entirely unscientific.
        \item Score 5 if the project description is factually accurate, well-researched, and scientifically sound.
    \end{itemize}
\end{itemize}

\subsection{Task 6: More case study}

\textbf{Task 6: Case Study}
To further understand how knowledge graphs could help the reasoning, we present a case study on task 6 in Tab.~\ref{tab:case_study}. The posed question incorporates five concepts (highlighted in blue), with the task being to formulate a feasible project proposal. Although LLaMA offers a substantial project description, its content and relevance to the highlighted concepts (marked in orange) are somewhat lacking. In contrast, GPT-4o not only references the queried concepts but also provides detailed insights (highlighted in purple) on their potential utility within the project, such as the role of \textit{neural question answering}. Lastly, with Graphusion constructed KG, the model provides a more comprehensive solution, elaborating on the concepts and introducing additional ones (highlighted in lavender) that come from the recovered concept graph, like \textit{dependency parsing} and \textit{event extraction}, while initially addressing the queried concepts. 

\begin{table*}[h]
\centering
\small
\begin{tabularx}{\textwidth}{lXX}
\toprule
\textbf{\textit{Question}} & \textit{I already know about \textcolor{bar_color1}{neural question answering, social media analysis, topic modeling, relation extraction, and word embedding variations}. In the domain of natural language processing, what potential project can I work on? Give me a possible idea. Show me the title and project description.} \\
\midrule
\textbf{LLaMA} & \textbf{Title: }Exploring the Impact of Emojis on Textual Emotions; \textbf{Description:} Emojis have become an integral part of online communication...The project will involve the following tasks: \textcolor{bar_color6}{ 1. Data Collection: We will collect a dataset of text messages...and their corresponding emotional labels (positive, negative, neutral). 2. Data Preprocessing: We will preprocess the data by removing stop words... }\\
\textbf{GPT-4o} & \textbf{Title:} Multi-faceted Analysis of Social Media Discussions on Climate Change; 
\textbf{Description:} This project aims to ...
Utilizing \textcolor{bar_color1}{neural question answering}, we can automatically address common inquiries people have about climate topics. \textcolor{bar_color1}{Social media analysis} will uncover trends, sentiment, and influential users shaping the conversation. \textcolor{bar_color2}{Topic modeling will identify underlying themes in these discussions} ...\\
\textbf{Ours} & \textbf{Title:} Enhanced Social Media Insights through Multilingual Aspect-Based Sentiment Analysis; \textbf{Description:}: This project aims to develop a comprehensive system for analyzing social media data to extract and understand user sentiments across multiple languages. By leveraging \textcolor{bar_color1}{neural question answering, topic modeling, and relation extraction,} the system will identify key aspects and sentiments from posts. \textcolor{bar_color3}{Dependency parsing and event extraction} will help in understanding syntactic and event-based relations... 
\textcolor{bar_color2}{Word embedding variations and multilingual word embeddings} will ensure accurate sentiment analysis across languages...\\
\bottomrule
\end{tabularx}
\caption{Case study on TutorQA Task 6: LLaMA, GPT-4o, and our pipeline with constructed KG from Graphusion (part). We highlight the \textcolor{bar_color1}{relevant concepts}, \textcolor{bar_color6}{irrelevant concepts},  \textcolor{bar_color2}{concept details} and \textcolor{bar_color3}{expanded relevant concepts}.}
\label{tab:case_study}
\vspace{-2mm}
\end{table*}

\clearpage
\noindent In Table~\ref{tab:app_case_study}, we present another example. Similarly to previous observations, LLaMA tends to provide much irrelevant information, whereas GPT-4o and our model offer more viable project proposals. Furthermore, our model demonstrates the ability to suggest detailed concepts and incorporate related concepts as extensions. 

\begin{table*}[h]
\centering
\small
\begin{tabularx}{\textwidth}{lXX}
\toprule
\textbf{\textit{Question}} & \textit{I already know about \textcolor{bar_color1}{neural question answering, recurrent neural networks and gated recurrent units}. In the domain of natural language processing, what potential project can I work on? Give me a possible idea. Show me the title and project description.} \\
\textbf{LLaMA2} & \textbf{Title:} Exploring the Use of Multimodal Fusion for Enhancing Question Answering Systems; \textbf{Description:} \textcolor{bar_color1}{Question answering} systems have been a crucial component of natural language processing research in recent years. While most question answering systems rely solely on textual information, there is a growing interest in incorporating multimodal information (such as images, videos, and audio) to improve the accuracy and robustness of these systems. In this project, we aim to explore the use of \textcolor{bar_color6}{multimodal fusion} for enhancing question answering systems.  The project will involve the following tasks:  1. Collecting and annotating a large dataset of multimodal question answering examples, where each example consists of a question, a set of candidate answers, and relevant multimodal information (such as images, videos, or audio)...[\texttt{Stops because of the length restriction}]\\
\textbf{GPT-4o} & \textbf{Title: }Sentiment Analysis and Trend Prediction in Social Media Data;
\textbf{Description:} Leverage the power of natural language processing to analyze and predict trends in social media sentiment. Your project will focus on collecting and processing large volumes of social media posts from various platforms. Using \textcolor{bar_color3}{advanced sentiment analysis techniques}, you will classify these posts into categories like positive, negative, and neutral. The challenge is to accurately interpret not just the explicit text, but also the implied and contextual sentiment. Furthermore, \textcolor{bar_color3}{apply time-series analysis to predict future sentiment trends}, which can be invaluable for businesses, political campaigns, and public relations. This project combines NLP with data science to provide actionable insights from the vast and dynamic world of social media. \\
\textbf{Ours} & \textbf{Title: }Adaptive Learning Assistant for Personalized Education; \textbf{Description: }The project aims to develop an Adaptive Learning Assistant (ALA) that leverages \textcolor{bar_color1}{neural question answering, recurrent neural networks (RNNs), and gated recurrent units (GRUs)} to provide personalized educational support. By integrating \textcolor{bar_color3}{text mining techniques}, the ALA will analyze educational content and student interactions to identify knowledge gaps and learning preferences. Using document ranking, it will curate and recommend personalized learning materials and quizzes. The system will employ \textcolor{bar_color3}{character-level language models} to understand and generate nuanced responses to student inquiries, facilitating a more interactive and engaging learning experience. Additionally, incorporating a \textcolor{bar_color3}{neural Turing machine} architecture will enable the ALA to remember and utilize past interactions to improve its recommendations over time, adapting to the evolving needs of each learner. This project not only aims to enhance the educational experience by providing tailored support but also seeks to explore the boundaries of current AI capabilities in understanding and responding to complex human learning behaviors.\\
\bottomrule
\end{tabularx}
\caption{Case study on TutorQA Task 6: We compare the results of LLaMA2, GPT-4o, and GPT-4-Graphusion. We highlight the \textcolor{bar_color1}{relevant concepts}, \textcolor{bar_color6}{irrelevant concepts},  \textcolor{bar_color2}{concept details} and \textcolor{bar_color3}{expanded relevant concepts}.}
\label{tab:app_case_study}
\end{table*}

\clearpage
\section{Knowledge Graph Construction Analysis}

\begin{figure*}[h]
    \centering
    \begin{adjustbox}{center}
    \begin{tikzpicture}
    \centering
    \small
    \begin{axis}[
        ybar,
        bar width=12pt,
        enlarge x limits=0.1,
        enlarge y limits={value=0.2,upper},
        legend style={
            at={(0.5,-0.2)}, 
            anchor=north,    
            legend columns=4 
        },
        ylabel={Average Rating},
        symbolic x coords={Is-a-Prerequisite-of, Used-for, Part-of, Hyponym-Of, Compare, Evaluate-for, Conjunction},
        xtick=data,
        height=8cm,
        width=18cm,
        nodes near coords,
        nodes near coords align={vertical},
        every node near coord/.append style={
            font=\footnotesize,
            /pgf/number format/.cd,
            fixed,
            precision=1,
            fixed zerofill
        }
    ]

    \addplot[fill=color1, draw=none] coordinates {(Is-a-Prerequisite-of, 2.9) (Used-for, 2.9) (Part-of, 2.8) (Hyponym-Of, 2.9) (Compare, 2.3) (Evaluate-for, 2.8) (Conjunction, 2.7)};
    \addlegendentry{LLaMA}

    \addplot[fill=color2, draw=none] coordinates {(Is-a-Prerequisite-of, 2.9) (Used-for, 3.0) (Part-of, 3.0) (Hyponym-Of, 0.0) (Compare, 2.8) (Evaluate-for, 3.0) (Conjunction, 1.0)};
    \addlegendentry{GPT-3}

    \addplot[fill=color3, draw=none] coordinates {(Is-a-Prerequisite-of, 2.9) (Used-for, 2.8) (Part-of, 3.0) (Hyponym-Of, 2.9) (Compare, 3.0) (Evaluate-for, 3.0) (Conjunction, 3.0)};
    \addlegendentry{GPT-4}

    \addplot[fill=color4, draw=none] coordinates {(Is-a-Prerequisite-of, 2.9) (Used-for, 2.9) (Part-of, 2.9) (Hyponym-Of, 3.0) (Compare, 3.0) (Evaluate-for, 3.0) (Conjunction, 2.3)};
    \addlegendentry{GPT-4o}

    \end{axis}
    \end{tikzpicture}
    \end{adjustbox}
    \caption{Relation entity quality rating by human evaluation, grouped by relation type.}
    \label{fig:relation_quality}
\end{figure*}
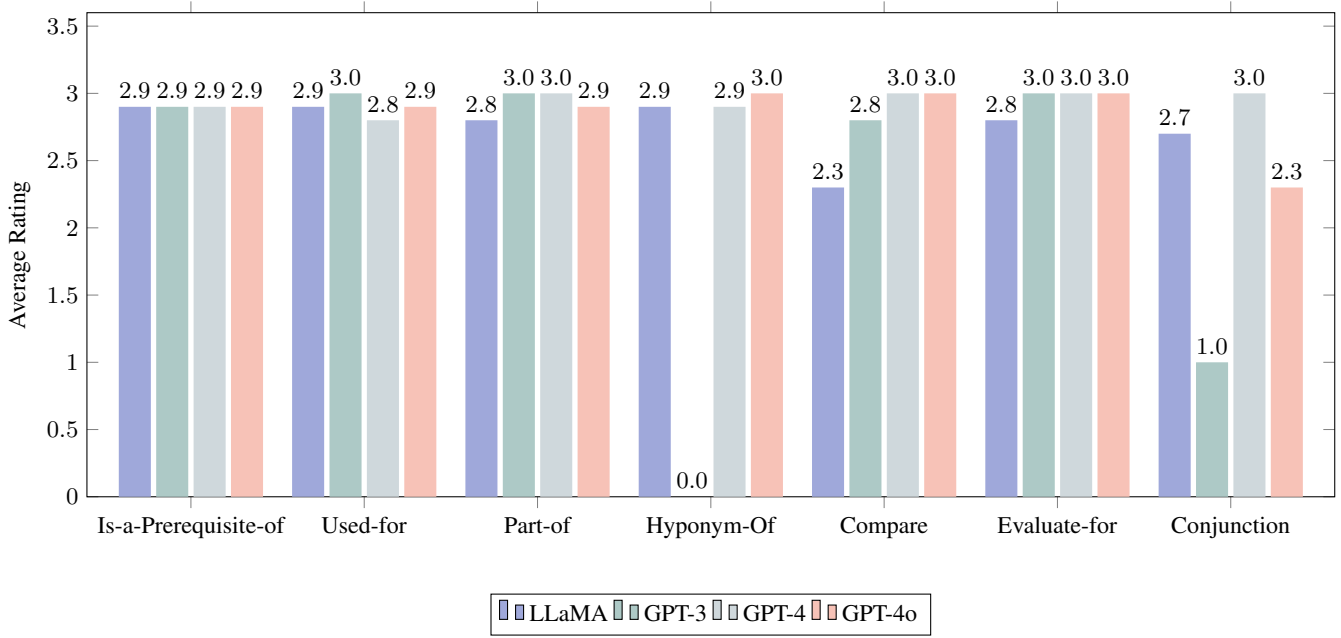

\textbf{Average Rating} We compare expert ratings on the Graphusion KGC results produced by four models: LLaMA, GPT-3.5, GPT-4, and GPT-4o. Fig.~\ref{fig:concept_quality} and \ref{fig:relation_quality} display the average ratings for concept entity quality and relation quality, respectively, grouped by relation type. Most types achieve an average rating of around 3 (full score) in concept entity quality, indicating that the extracted triplets contain good in-domain concepts. In contrast, the ratings for relation quality are slightly lower. GPT-4 and GPT-4o perform better in relation prediction.

\begin{figure*}[h]
    \centering
    \begin{adjustbox}{center}
    \begin{tikzpicture}
    \centering
    \small
    \begin{axis}[
        ybar,
        bar width=12pt,
        enlarge x limits=0.1,
        enlarge y limits={value=0.2,upper},
        legend style={
            at={(0.5,-0.2)}, 
            anchor=north,    
            legend columns=4 
        },
        ylabel={Average Rating},
        symbolic x coords={Is-a-Prerequisite-of, Used-for, Part-of, Hyponym-Of, Compare, Evaluate-for, Conjunction},
        xtick=data,
        height=8cm,
        width=18cm,
        nodes near coords,
        nodes near coords align={vertical},
        every node near coord/.append style={
            font=\footnotesize,
            /pgf/number format/.cd,
            fixed,
            precision=1,
            fixed zerofill
        }
    ]

    \addplot[fill=color1, draw=none] coordinates {(Is-a-Prerequisite-of, 2.9) (Used-for, 2.9) (Part-of, 2.8) (Hyponym-Of, 2.9) (Compare, 2.3) (Evaluate-for, 2.8) (Conjunction, 2.7)};
    \addlegendentry{LLaMA}

    \addplot[fill=color2, draw=none] coordinates {(Is-a-Prerequisite-of, 2.9) (Used-for, 3.0) (Part-of, 3.0) (Hyponym-Of, 0.0) (Compare, 2.8) (Evaluate-for, 3.0) (Conjunction, 1.0)};
    \addlegendentry{GPT-3}

    \addplot[fill=color3, draw=none] coordinates {(Is-a-Prerequisite-of, 2.9) (Used-for, 2.8) (Part-of, 3.0) (Hyponym-Of, 2.9) (Compare, 3.0) (Evaluate-for, 3.0) (Conjunction, 3.0)};
    \addlegendentry{GPT-4}

    \addplot[fill=color4, draw=none] coordinates {(Is-a-Prerequisite-of, 2.9) (Used-for, 2.9) (Part-of, 2.9) (Hyponym-Of, 3.0) (Compare, 3.0) (Evaluate-for, 3.0) (Conjunction, 2.3)};
    \addlegendentry{GPT-4o}

    \end{axis}
    \end{tikzpicture}
    \end{adjustbox}
    \caption{Concept entity quality rating by human evaluation, grouped by relation type.}
    \label{fig:concept_quality}
\end{figure*}
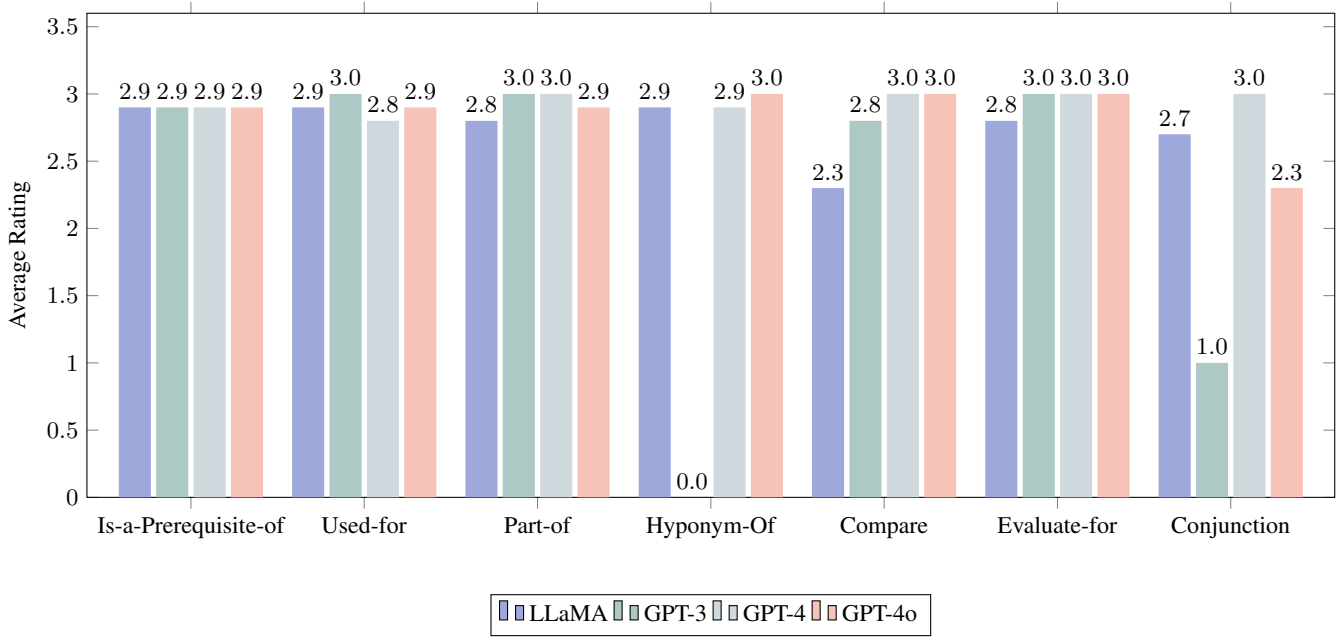

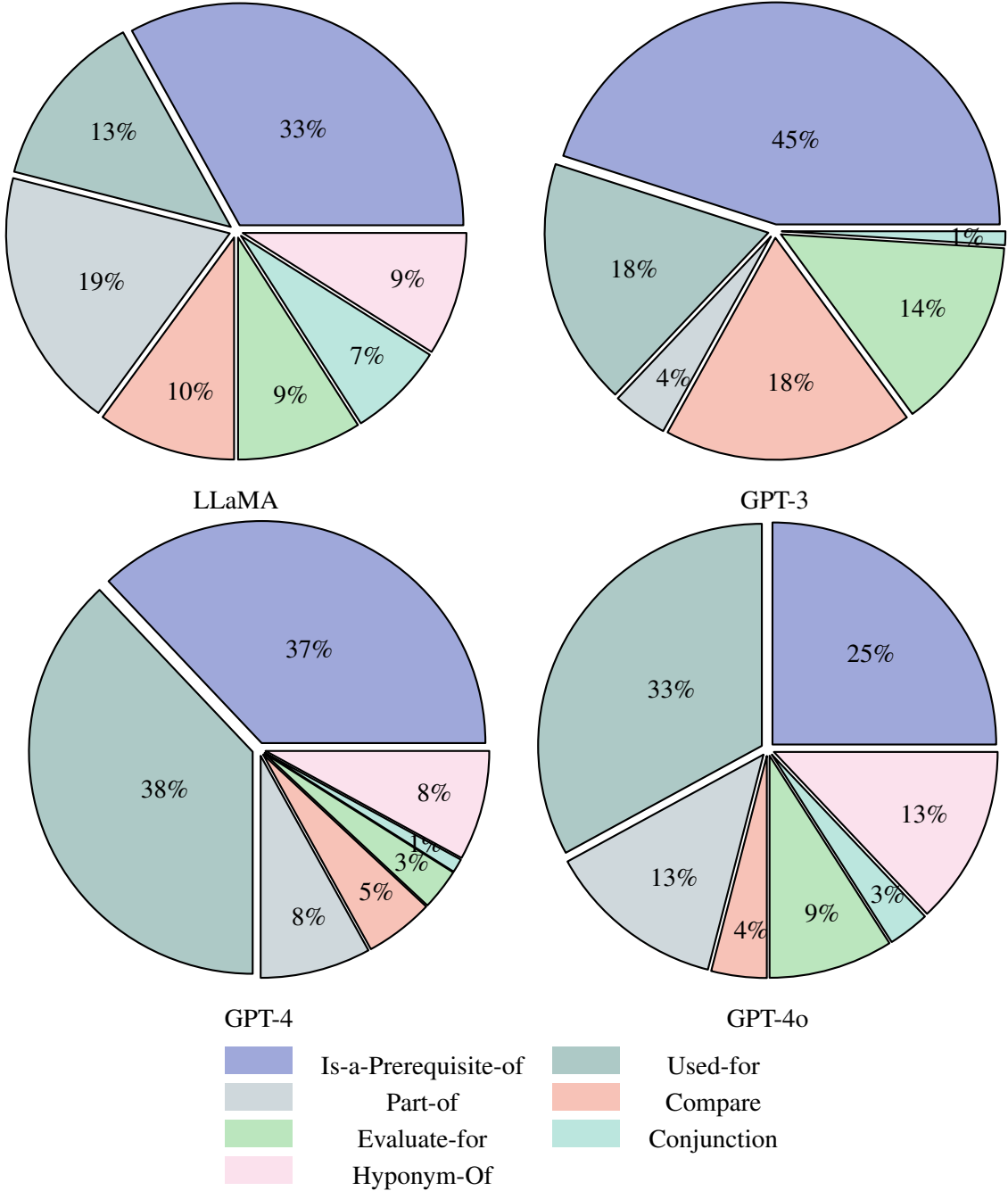
\begin{figure*}[h]
    \centering
    \begin{tikzpicture}
        \pie[color={color1,color2,color3,color4,color5,color6,pie_rose},
             text=,
             radius=3.3,
             before number=\phantom{0},
             after number={\%\scriptsize},
             explode=0.1]{
            33/Is-a-Prerequisite-of,
            13/Used-for,
            19/Part-of,
            10/Compare,
            9/Evaluate-for,
            7/Conjunction,
            9/Hyponym-Of
        }
        \node at (0, -4) {LLaMA};
    \end{tikzpicture}
    \hspace{0.7cm}
    \begin{tikzpicture}
        \pie[color={color1,color2,color3,color4,color5,color6,pie_rose},
             text=,
             radius=3.3,
             before number=\phantom{0},
             after number={\%\scriptsize},
             explode=0.1]{
            45/Is-a-Prerequisite-of,
            18/Used-for,
            4/Part-of,
            18/Compare,
            14/Evaluate-for,
            1/Conjunction
        }
        \node at (0, -4) {GPT-3};
    \end{tikzpicture}
    \hspace{0.7cm}
    \begin{tikzpicture}
        \pie[color={color1,color2,color3,color4,color5,color6,pie_rose},
             text=,
             radius=3.3,
             before number=\phantom{0},
             after number={\%\scriptsize},
             explode=0.1]{
            37/Is-a-Prerequisite-of,
            38/Used-for,
            8/Part-of,
            5/Compare,
            3/Evaluate-for,
            1/Conjunction,
            8/Hyponym-Of
        }
        \node at (0, -4) {GPT-4};
    \end{tikzpicture}
    \hspace{0.5cm}
    \begin{tikzpicture}
        \pie[color={color1,color2,color3,color4,color5,color6,pie_rose},
             text=,
             radius=3.3,
             before number=\phantom{0},
             after number={\%\scriptsize},
             explode=0.1]{
            25/Is-a-Prerequisite-of,
            33/Used-for,
            13/Part-of,
            4/Compare,
            9/Evaluate-for,
            3/Conjunction,
            13/Hyponym-Of
        }
        \node at (0, -4) {GPT-4o};
    \end{tikzpicture}
    \begin{tikzpicture}
        \node at (0, 0) {\begin{tabular}{cccc}
            \textcolor{color1}{\rule{1cm}{0.4cm}} & Is-a-Prerequisite-of &
            \textcolor{color2}{\rule{1cm}{0.4cm}} & Used-for \\
            \textcolor{color3}{\rule{1cm}{0.4cm}} & Part-of &
            \textcolor{color4}{\rule{1cm}{0.4cm}} & Compare \\
            \textcolor{color5}{\rule{1cm}{0.4cm}} & Evaluate-for &
            \textcolor{color6}{\rule{1cm}{0.4cm}} & Conjunction \\
            \textcolor{pie_rose}{\rule{1cm}{0.4cm}} & Hyponym-Of & & \\
        \end{tabular}};
    \end{tikzpicture}
    \caption{Relation type distribution.}
    \label{fig:relation type distribution}
\end{figure*}

\clearpage
\noindent\textbf{Relation Type Distribution}  
We then compare the Graphusion results for each relation type across the four selected base LLMs, as shown in Fig.~\ref{fig:relation type distribution}. All models tend to predict \texttt{Prerequisite\_of} and \texttt{Used\_For} relations. The results from LLaMA show relatively even distributions across relation types, whereas the results from the GPT family do not.

\begin{figure}[h]
    \centering
\includegraphics[width=1\textwidth]{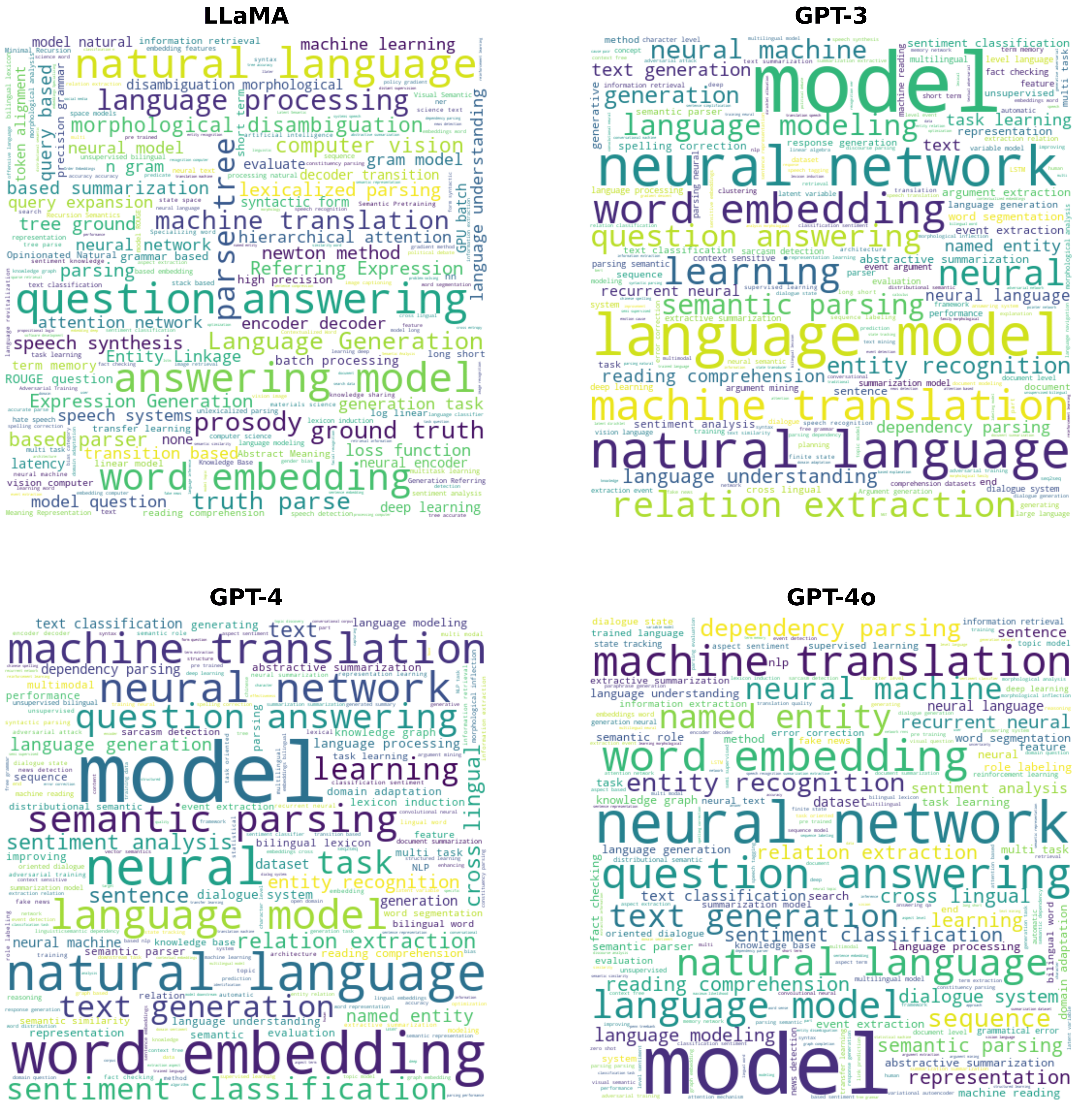}
\vspace{-2mm}
    \caption{Word cloud visualization for extracted concepts.}
    \label{fig:word_cloud}
    \vspace{-3mm}
\end{figure}

\clearpage
\noindent\textbf{Word cloud Visualization} Finally, in Fig.~\ref{fig:word_cloud}, we present a word cloud visualization of the concepts extracted by Graphusion, comparing the four base LLMs. High-frequency concepts include \texttt{word embedding}, \texttt{model}, \texttt{neural network}, \texttt{language model}, and others.

\end{document}